\pdfoutput=1
\documentclass{article}

\usepackage{iclr2027_conference_arxiv,times}
\usepackage[T1]{fontenc}
\usepackage[utf8]{inputenc}
\usepackage{microtype}
\usepackage{amsmath,amssymb,amsfonts,mathtools}
\usepackage{amsthm}
\newtheorem{theorem}{Theorem}
\newtheorem{lemma}{Lemma}
\newtheorem{proposition}{Proposition}

\usepackage{booktabs,tabularx,array}
\usepackage{graphicx}
\usepackage{xcolor}
\usepackage{hyperref}
\usepackage{enumitem}
\usepackage{algorithm}
\usepackage{algpseudocode}
\usepackage{caption}
\usepackage{subcaption}
\usepackage{placeins}
\usepackage{natbib}
\usepackage{url}
\usepackage{xspace}

\hypersetup{colorlinks=true, linkcolor=blue!50!black, citecolor=blue!50!black, urlcolor=blue!50!black}
\setlist[itemize]{leftmargin=*,topsep=3pt,itemsep=2pt,parsep=0pt}
\setlist[enumerate]{leftmargin=*,topsep=3pt,itemsep=2pt,parsep=0pt}
\algrenewcommand\algorithmicrequire{\textbf{Input:}}
\algrenewcommand\algorithmicensure{\textbf{Output:}}
\algrenewcommand\alglinenumber[1]{\scriptsize #1:}
\title{CineForge: Self-Improving Agents for Long-Horizon Video Generation}
\author{
{\normalfont\bfseries Junxiang Liu$^{1,*}$ \quad
Lin Wang$^{2,*}$ \quad
Haiyu Shi$^{2}$ \quad
Hongxu Ma$^{3}$ \quad
Xiaoyu Yang$^{2}$} \\
{\normalfont\bfseries Chunjie Chen$^{2}$ \quad
Xiaoxiao Xu$^{2}$ \quad
Kaiqiao Zhan$^{2}$ \quad
Boao Wang$^{1}$} \\
{\normalfont\bfseries Shuizhou Shi$^{1}$ \quad
Tianyun Zhu$^{1}$ \quad
Jie Li$^{1}$ \quad
Jiangtong Li$^{1}$} \\
{\normalfont\small $^{1}$Tongji University, Shanghai, China \qquad
$^{2}$Kuaishou Technology, Beijing, China} \\
{\normalfont\small $^{3}$Fudan University, Shanghai, China} \\
{\normalfont\footnotesize\texttt{\{junxiang\_liu,2534143,2534170,Zhutianyun,jieli,jiangtongli\}@tongji.edu.cn}} \\
{\normalfont\footnotesize\texttt{\{wanglin36,shihaiyu,yangxiaoyu,xuxiaoxiao05,zhankaiqiao\}@kuaishou.com}} \\
{\normalfont\footnotesize\texttt{hxma24@m.fudan.edu.cn \quad chencj517@gmail.com}}
}
\date{}

\begin{document}
\maketitle
\begingroup
\renewcommand{\thefootnote}{*}
\footnotetext[1]{Equal contribution.}
\endgroup

\begin{abstract}
Long-horizon story-driven video generation requires a production agent to coordinate narrative decomposition, state tracking, shot design, prompt construction, rendering, and revision across interdependent scenes. 
Existing adaptive video systems primarily refine requests or reusable skills, leaving recurring production failures disconnected from persistent, stage-targeted improvements across stories. 
We introduce \textbf{CineForge}, a self-evolving video-production agent framework that couples \textbf{CineForge-Produce} for video generation with \textbf{CineForge-Evolve} for cross-story policy evolution. 
CineForge-Produce organizes each source story into typed narrative, character, spatial, and cinematic states, uses them to coordinate asset and clip generation, and records the process as a canonical production trajectory. 
CineForge-Evolve applies \textbf{Case-to-Pattern-to-Policy Evolution (CPPE)} to review trajectory evidence, consolidate recurrent findings into bounded stage-local patches, and deploy validated updates through structural replay and confidence-controlled paired evaluation. 
To measure complete story realization, we introduce \textbf{CineScope}, which combines a 100-script \textbf{CineScope-Data} suite with a human-aligned, multiscale \textbf{CineScope-Metric} spanning causal state, directorial orchestration, pacing and resource allocation, and character arc. 
Across CineScope-Data and two public benchmarks, the evolved CineForge policy improves CineScope-Metric from 4.024 to 4.380, outperforms three long-video baselines with consistent gains under ScriptAgent, and reduces review LLM calls by 37.0\% on new stories. These results establish production trajectories as actionable experience for video agents that improve cumulatively across long-form storytelling tasks.
\end{abstract}

\section{Introduction}

Recent advances in video foundation models have substantially improved visual fidelity, motion realism, instruction following, and temporal extension \citep{kim2024fifo,lu2024freelong,guo2025longcontext,kara2025shotadapter}. These capabilities are moving video generation toward long-form story-driven production, a long-horizon generation setting in which a complete narrative is realized through a sequence of interdependent scenes and shots. Success in this setting depends on preserving event coverage, character identity and state, spatial relations, directorial intent, and pacing throughout the full narrative horizon. This paper studies how a persistent video-production agent can learn from completed productions and convert prior experience into systematic improvements for future stories.

\begin{figure}[!t]
    \centering
    \includegraphics[width=\textwidth]{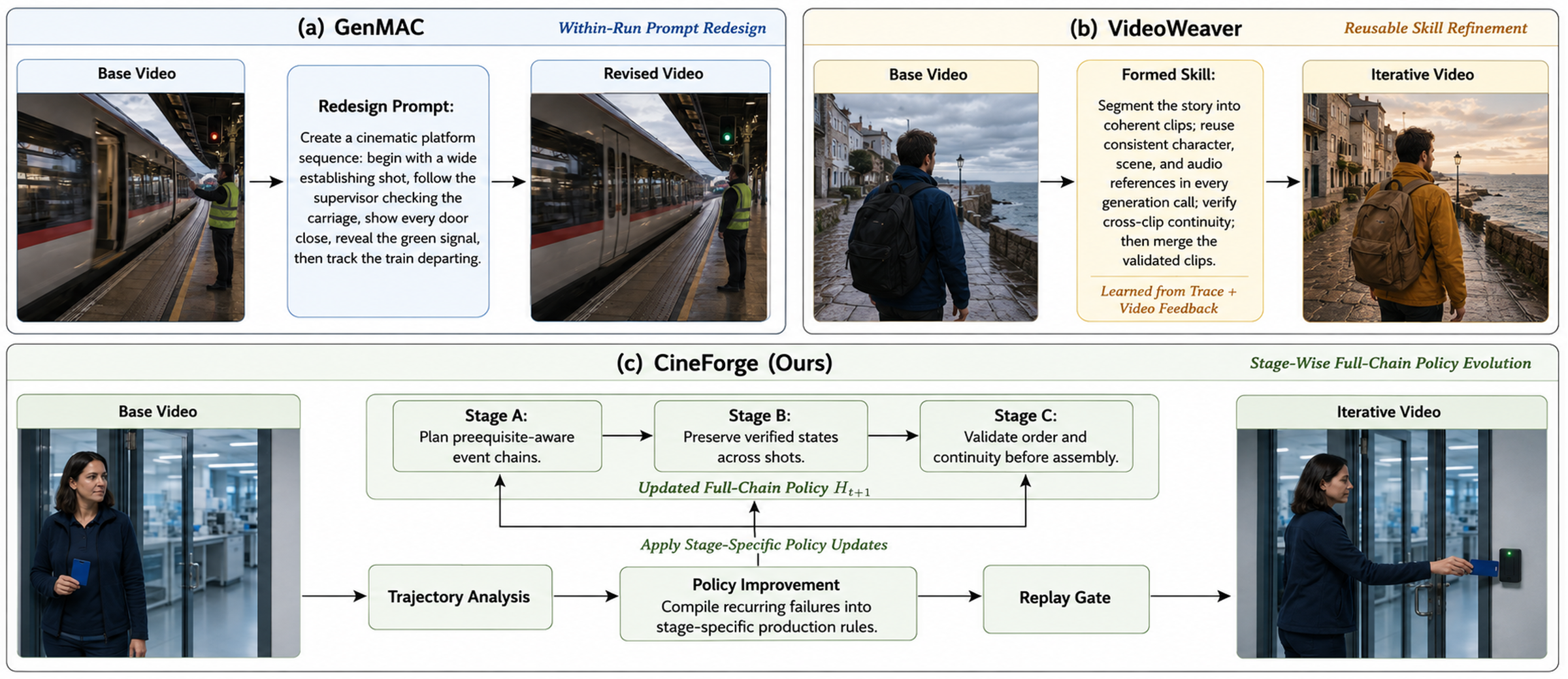}
    \caption{Comparison of adaptation scopes: instance-level repair (GenMAC), skill-level evolution (VideoWeaver), and our trajectory-level replay-gated policy evolution (CineForge).}
    \label{fig:method-comparison}
\end{figure}

Agentic video systems increasingly combine hierarchical planning, memory, multi-agent coordination, iterative critique, and consistency control to manage complex long-horizon generation workflows \citep{lin2023videodirectorgpt,wang2024aesopagent,wu2025movieagent,long2026a2rd}. Recent adaptive approaches extend this direction across several distinct adaptation scopes: GenMAC iteratively repairs the current generation, while VideoWeaver evolves reusable skills from prior generated videos \citep{huang2024genmac,wei2026videoweaver}. Figure~\ref{fig:method-comparison} positions these methods as instance-level and skill-level adaptation, respectively. Long-form production calls for a further trajectory-level capability: persistent production-policy evolution across stories, since a visible defect may originate in narrative decomposition, state tracking, shot design, prompt rendering, or backend execution. Realizing this capability requires production-stage traceability, cross-case recurrence modeling, stage-targeted policy updates, and deployment gates that preserve reliable behavior.

We introduce \textbf{CineForge}, a self-evolving agent framework for long-horizon story-driven video production. CineForge organizes the full production lifecycle into two coupled components: \textbf{CineForge-Produce} and \textbf{CineForge-Evolve}. CineForge-Produce transforms a source story into typed narrative and cinematic states, generates the corresponding assets, clips, and composed episode, and records the complete production trajectory. CineForge-Evolve converts accumulated production experience into versioned policy improvements that guide subsequent stories, establishing a closed loop from video production to persistent agent evolution.

CineForge-Produce first decomposes each source story into checkable narrative units and organizes them into episode, scene, shot, character, and spatial states. These shared states coordinate asset creation, prompt construction, clip generation, composition, and bounded revision, while the agent records intermediate decisions and outcomes as a canonical production trajectory. CineForge-Evolve applies \textbf{Case-to-Pattern-to-Policy Evolution (CPPE)} to review pipeline artifacts, local clips, and complete episodes, linking observed failures to the earliest production stages supported by the trajectory evidence. Across stories, recurrent findings are consolidated into patterns and compiled into bounded policy patches that target the corresponding production stages. Each candidate patch is replayed against the incumbent policy through structural checks and confidence-controlled paired evaluations, allowing validated production knowledge to enter the next policy version.

Reliable progress in long-form video production requires an evaluation framework that captures story-level quality across episodes. We therefore introduce \textbf{CineScope}, comprising \textbf{CineScope-Data} and \textbf{CineScope-Metric}. CineScope-Data contains 100 long-form benchmark scripts curated across diverse genres, narrative structures, character configurations, temporal spans, and scene-transition patterns, providing evaluation inputs that exercise long-range story dependencies. CineScope-Metric evaluates episodes through multiscale video evidence and twenty human-aligned criteria grouped into Causal State, Directorial Orchestration, Pacing and Resource Allocation, and Character Arc. Calibrated against expert judgments and frozen for system evaluation, CineScope provides a foundation for measuring story realization quality and tracking improvements across evolution rounds.

Our experiments follow a two-stage evaluation protocol. We first establish CineScope-Metric as a reliable measure of story-level video quality by comparing its assessments with expert judgments on complete episodes. We then freeze the metric and evaluate CineForge on CineScope-Data and two public benchmarks, AnimeShooter and ViStoryBench \citep{qiu2025animeshooter,vistorybench2025}, with ScriptAgent providing an additional story-level view and VBench-Long measuring complementary perceptual properties. CineScope-Metric achieves an overall Spearman correlation of 0.710 and a Bradley-Terry (BT) pair agreement of 0.810 with expert judgments. These scores substantially exceed the 0.4--0.5 range at which automated metrics often plateau on complex video tasks, establishing CineScope-Metric as a highly reliable proxy for human evaluation and providing a robust basis for subsequent comparisons. Under this validated protocol, the evolved CineForge policy reaches 4.380, an 8.8\% relative improvement over Round~0 (4.024), and outperforms three long-video baselines under CineScope-Metric, with consistent gains under ScriptAgent. Because agent-based video generation often plateaus once a strong Round~0 policy has been established, and further gains require resolving complex long-range narrative dependencies, this margin represents a substantial improvement in story-level coherence rather than a marginal refinement. Successive fresh evolution rounds produce sustained improvements, with Round~3 gaining 0.355 in CineScope and 0.188 in ScriptAgent over Round~0. On new stories, the evolved agent improves both outcome scores while, by Round~1, reducing review LLM calls by 23.7\% and detected review issues by 30.8\%. Crucially, these reductions translate directly into computational savings and a higher first-pass yield in practical pipelines, demonstrating that our method not only generates better videos but also does so with significantly lower inference overhead. Overall, these results demonstrate cumulative, transferable, and more efficient production-policy evolution. Our contributions are as follows:
\begin{itemize}[leftmargin=*,itemsep=1pt,topsep=2pt]
    \item We introduce CineForge, which unifies CineForge-Produce and CineForge-Evolve to connect structured long-horizon video production with persistent policy improvement across stories.
    \item We develop CPPE to transform production-trajectory findings into recurrent patterns and bounded stage-local patches through replay-gated deployment and confidence-controlled analysis.
    \item We introduce CineScope, combining a 100-script CineScope-Data suite with the human-aligned, multiscale CineScope-Metric for evaluating complete story-driven videos.
    \item Experiments across CineScope and public benchmarks demonstrate strong video quality, cumulative evolution gains, transfer to new stories, and reduced production-review workload.
\end{itemize}

\section{Preliminaries}
\label{sec:formulation}

This section defines the production objects, update units, and evaluation boundaries of our framework.

\subsection{Task, Production Trace, and Persistent Policy}

Let $r$ index a policy version and its collection round, and let $i$ index a story in that round. For a long story or script-like task $x_{r,i}\sim\mathcal{D}$, CineForge-Produce executes the persistent policy $H_r$ under generation randomness $\zeta_{r,i}$ and returns $(V_{r,i},Z_{r,i})=\operatorname{CineForge\text{-}Produce}(x_{r,i};H_r,\zeta_{r,i}),$ where $V_{r,i}$ is the story video and $Z_{r,i}$ is its canonical production trajectory. 
$\mathcal D$ is the target task distribution, with case collection, fresh admission, and held-out evaluation using their protocol-specific samples. The typed trajectory records stage inputs and outputs, validator and Review-Rewrite outcomes, provenance, and dependency links across narrative atoms, episode, scene, and shot plans, character and spatial states, prompts, assets, and generated clips. 

The persistent production policy, or harness, is $H_r=(\mathcal{C}_{stage},\mathcal{R}_{prompt},\Theta_{\mathrm{prod}},\mathcal{R}_{review},\mathcal{P}_{style}),$ whose components denote stage contracts, prompt-rendering rules, production thresholds and retry budgets, review-routing rules, and bounded style profiles. Recurrence thresholds, replay construction, confidence budgets, and acceptance margins govern evolution as fixed protocol controls within a certified sequence. CineForge-Produce realizes $H_r$ in the current production, bounded in-run repair revises current artifacts and records the outcome in $Z_{r,i}$, and an accepted CineForge-Evolve patch creates $H_{r+1}=H_r\oplus\Delta_r^\star$ for subsequent stories.

\subsection{Cases, Patterns, and Bounded Updates}

A finding records a production symptom as $\ell=(d,u,o,\mathcal E,\mathcal C_0),$ where $d$ identifies the discovery stream, such as pipeline, local-video, or global-video review; $u$ is the observation unit, such as an episode, scene, shot, prompt, or clip; $o$ is the symptom, $\mathcal E$ is the evidence, and $\mathcal C_0$ is the candidate-cause set. Review and diagnosis normalize the finding into a standard issue $e=(\sigma,u,g,o,\mathcal E,f,m),$ where $\sigma$ is the earliest evidence-supported causal stage, $g$ is the normalized failure family, $f$ is the proposed repair specification, and $m$ contains confidence, lineage, operational context, and repair status. Thus, $\ell$ records what was observed, while $e$ records the diagnosis and its response. 

The normalized issues from one production form a production case $\kappa_{r,i}=(x_{r,i},H_r,Z_{r,i},V_{r,i},\mathcal I_{r,i},\mathcal A_{r,i}),$ where $\mathcal I_{r,i}$ is the standard-issue set and $\mathcal A_{r,i}$ contains Review-Rewrite actions and outcomes. Validator records, evidence, and provenance remain linked through $Z_{r,i}$. In Appendix~\ref{app:method}, we specify the complete case schema.

A recurring pattern is $P=(\sigma,g,\chi,E_P),$ where $\chi$ summarizes the conditions under which the failure recurs and $E_P$ is the supporting production-case set. Each production contributes one case, so eligibility requires a predeclared threshold $|E_P|\ge k_g$ over distinct story-level opportunities supported by sufficiently confident, policy-editable issues. An eligible pattern produces a bounded, stage-local policy patch $\Delta=(\texttt{target},\texttt{edit\_type},\texttt{payload},\rho,\texttt{rollback})$ and $H'=H\oplus\Delta$, where $\rho$ is the risk level. The typed target belongs to an editable $H$ component, the edit type restricts allowed operations, the payload contains the change, and rollback metadata identifies the incumbent version.

\subsection{Replay Regimes and Evaluation Boundary}

Every candidate patch undergoes deterministic structural replay, while a patch affecting stochastic generation additionally undergoes paired stochastic evaluation. At round $r$, $R_r^{\mathrm{det}}$ contains fixed replay items evaluated from a declared boundary $b$. The replay context $c_b(x)$ stores fixed upstream artifacts, task inputs, and validator versions, while editable prompts, templates, and rules remain in $H_r$. When every affected transformation after $b$ is deterministic, $T_{H_r}^{d}(x;c_b)$ denotes the resulting trace and protected validators are compared exactly between $H_r$ and $H_r\oplus\Delta$. A patch that invokes a new LLM, image, or video-generation call is also evaluated on the paired stochastic set $R_r^{\mathrm{vis}}$ through repeated generation and scoring.

The risk-adjusted structural margin $Y_{H_r,\Delta}$ combines aggregate structural validity $Q$, protected-validator regression $D$, and changes outside the audited dependency slice $U$; their exact definitions are given in Appendix~\ref{app:proofs}. Final-video quality is measured after the evaluated policy version is fixed. For a frozen outcome evaluator $j$, we report $J_j(H)=\mathbb{E}_{x\sim\mathcal D,\zeta}[j(V_H(x;\zeta))]$ on protocol-defined held-out outputs, with CineScope-Metric and ScriptAgent as evaluator instances. Replay admission determines whether a patch may create the next policy version, while $J_j$ measures the rendered outcomes of that version.

\section{CineForge Framework}
\label{sec:method}

\subsection{Overview}

\begin{figure}[!t]
    \centering
    \includegraphics[width=\textwidth]{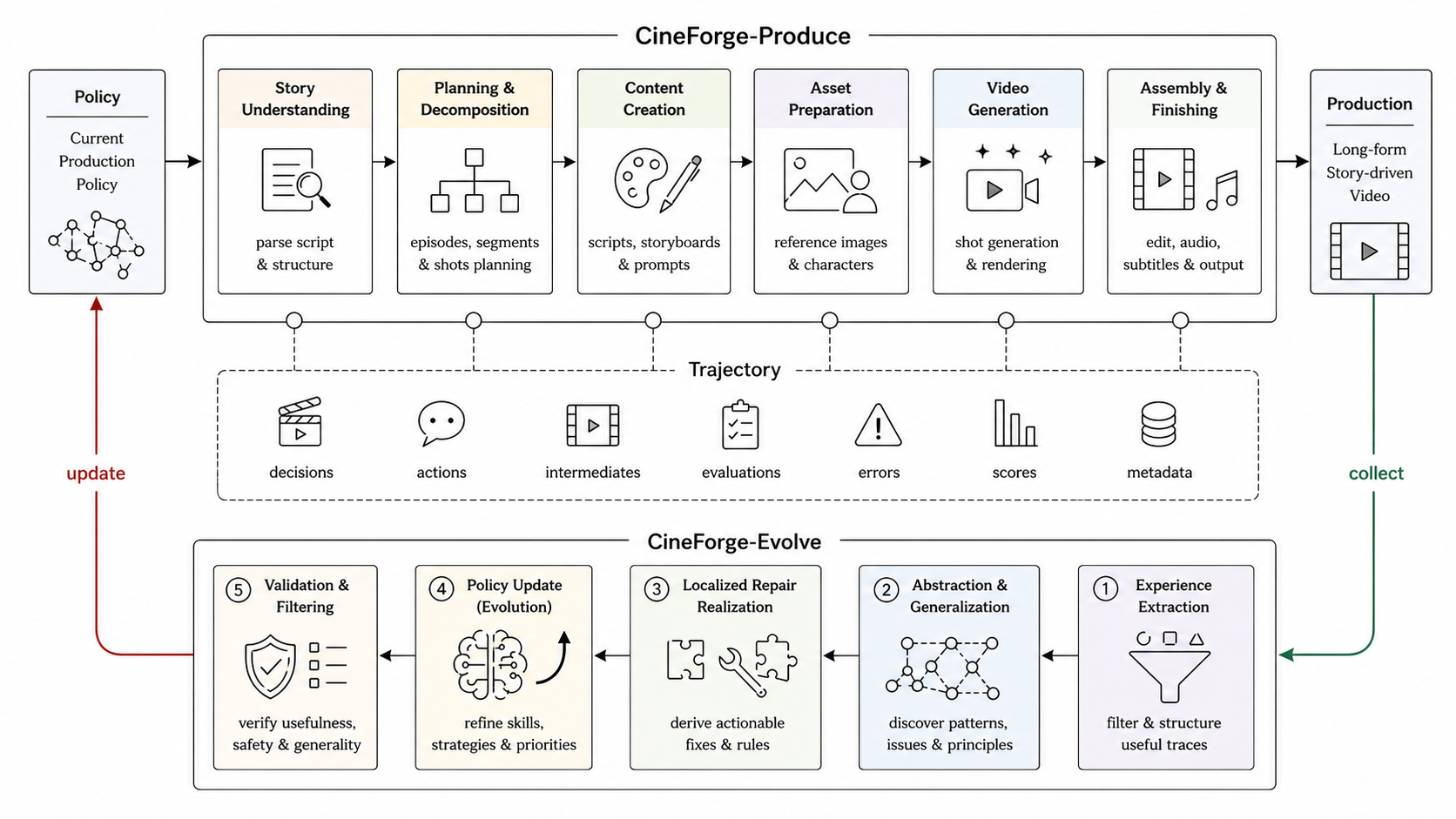}
    \caption{CineForge-Produce generates story videos and canonical production trajectories under the current policy. CineForge-Evolve converts recurring trajectory evidence into bounded, replay-validated policy updates for subsequent productions.}
    \label{fig:system-overview}
\end{figure}

CineForge couples trajectory-aware long-form production with cross-story policy evolution, as shown in Figure~\ref{fig:system-overview}. \textbf{CineForge-Produce} executes policy $H_r$ to generate story videos and canonical production trajectories. \textbf{CineForge-Evolve} applies Case-to-Pattern-to-Policy Evolution (CPPE) to aggregate trajectory-grounded experience, compile bounded policy patches, and validate the next policy version. For policy version and collection round $r$ and story index $i$, their interface is
\[
\begin{aligned}
(V_{r,i},Z_{r,i})&=\operatorname{CineForge\text{-}Produce}(x_{r,i};H_r,\zeta_{r,i}),\\
D_r&=\{(x_{r,i},Z_{r,i},V_{r,i})\}_{i=1}^{n_{D,r}},\\
(\mathcal B_r,\mathcal C_r,\Delta_r^\star,M_{r+1})
&=\operatorname{CineForge\text{-}Evolve}\!\left(H_r;D_r,M_r, R_r^{\mathrm{det}},R_r^{\mathrm{vis}}\right),\\
H_{r+1}&=\begin{cases}H_r\oplus\Delta_r^\star,&\Delta_r^\star\ne\varnothing,\\H_r,&\Delta_r^\star=\varnothing, \end{cases}
\end{aligned}
\]
where $D_r$ is the collection batch, $M_r$ is CaseMemory, $\mathcal B_r$ is the recurring-pattern set, and $\mathcal C_r$ is the candidate-patch set. The replay pools are accessed after $\mathcal C_r$ is frozen. CineForge-Produce makes each production observable, while CineForge-Evolve accumulates cross-story evidence and deploys an admitted patch as the next policy version.

The persistent policy $H_r$ is the shared versioned control layer defined in Section~\ref{sec:formulation}. Each accepted patch targets an editable policy component, records rollback metadata, and enters subsequent production through $H_{r+1}$. Recurrence and admission controls remain fixed within each certified sequence.

\subsection{CineForge-Produce: Trajectory-Aware Long-Form Production}

Given a story $x$, CineForge-Produce constructs a typed production state $\Sigma$ by decomposing source obligations into narrative atoms organized across episode, scene, and shot units with narrative, character, spatial, cinematic, and prompt fields. Coverage maps link atoms to shots, while character and scene ledgers maintain identity, position, facing, gaze, visual anchors, and state transitions across scenes. These dependencies support coverage and continuity checks before rendering.

Using $\Sigma$ and $H_r$, the production path coordinates reusable assets and references, provider-specific prompts, clip generation, visual-anchor propagation, and final composition. Deterministic contracts validate schemas, identifiers, atom coverage, state consistency, durations, and provider constraints, while bounded Review-Rewrite repairs affected artifacts and records the results. CineForge-Produce returns the video $V_{r,i}$ with the canonical trajectory $Z_{r,i}$, which retains stage inputs and outputs, prompts, asset and clip identifiers, validator results, repairs, provenance, and dependency links for locating the earliest evidence-supported production stage. Appendix~\ref{app:method} details state and trajectory construction, and Appendix~\ref{app:algorithms} gives the production procedure.

\subsection{CineForge-Evolve: Case-to-Pattern-to-Policy Evolution}

\paragraph{Trajectory-grounded case formation.}
CineForge-Evolve begins from the completed video $V_{r,i}$ and trajectory $Z_{r,i}$, gathering evidence through three complementary streams. Pipeline review checks contracts, semantic propagation, shared assets, and cross-segment state; local-video review jointly inspects segment prompts and clips; and global-video review compares the adapted script with the complete episode. Each finding records its observation unit, symptom, evidence, and candidate upstream causes. CPPE then follows the trajectory dependencies from the observed artifact through clip generation, prompt rendering, shot design, script, assets, and episode planning to assign the earliest evidence-supported causal stage $\sigma$. Ambiguous evidence is labeled \texttt{unknown}.

Findings across granularities are deduplicated and merged before a stage-specific critic normalizes the failure family $g$, verifies the causal stage, and proposes a bounded repair specification $f$ with confidence and lineage. The resulting standard issues form the production case $\kappa_{r,i}=(x_{r,i},H_r,Z_{r,i},V_{r,i},\mathcal I_{r,i},\mathcal A_{r,i})$, where $\mathcal A_{r,i}$ records Review-Rewrite actions and outcomes. CaseMemory retains the complete evidence chain, while sufficiently confident, policy-editable issues that remain unresolved or \texttt{stuck} contribute to cross-story pattern induction. Repair outcomes provide additional evidence about whether a failure persists across productions.

\paragraph{Cross-story abstraction and bounded patch compilation.}
CaseMemory aggregates eligible issues across stories, and CPPE groups their production cases by evidence-supported stage $\sigma$, normalized failure family $g$, and context signature $\chi$. Each group forms a recurring pattern $P=(\sigma,g,\chi,E_P)$ with supporting case set $E_P$. A pattern enters $\mathcal B_r$ when $|E_P|\ge k_g$ over distinct story-level opportunities, turning repeated trajectory evidence into a candidate for persistent improvement. The predeclared threshold $k_g$ and opportunity definition remain fixed within a certified sequence; Lemma~\ref{thm:recurrence} characterizes recurrence reliability.

For each $P\in\mathcal B_r$, CPPE maps the diagnosed failure and repair specification to a finite set of typed, stage-local patches; their union forms $\mathcal C_r$. A patch may refine a stage contract, add a required field, adjust a production threshold or review route, strengthen a rendering rule, or modify a bounded style profile. Missing narrative atoms, for example, induce coverage constraints, while recurrent spatial inconsistencies induce ledger rules. Each patch records its target, edit type, payload, risk level, dependency slice, and rollback metadata. CPPE freezes $\mathcal C_r$, the selection rule, and the tie-breaker before fresh admission, providing a fixed candidate class for replay.

\paragraph{Replay-gated policy versioning.}
Replay evaluates the fixed candidate class before policy versioning. Motivating failures and near misses form $R_r^{\mathrm{diag}}$; after $\mathcal C_r$ is frozen, CineForge-Evolve draws a fresh set $R_r$ for admission. Every candidate is compared with $H_r$ on $R_r^{\mathrm{det}}=R_r^{\mathrm{diag}}\cup R_r$ using the protected deterministic guard metrics $\mathcal G_\rho$ for risk level $\rho$:
\[
S_j(H_r\oplus\Delta;x)\ge S_j(H_r;x)-\epsilon_{j,\rho}
\quad \forall x\in R_r^{\mathrm{det}},\ \forall S_j\in\mathcal G_\rho,
\]
where tolerances $\epsilon_{j,\rho}\ge0$ are predeclared and zero for low-risk structural updates. The gate checks target behavior and protected guards on replay items, while the fresh subset $R_r$ supplies the structural margin certified in Section~\ref{sec:analysis}. Appendix~\ref{app:proofs} defines the guard weights and structural welfare $Q$.

Patches that invoke a new LLM, image, or video-generation call additionally undergo repeated paired generation and evaluation on $R_r^{\mathrm{vis}}$. With $K_r=|\mathcal C_r|$, $n_{V,r}$ paired stories, admission-score weight sum $W_{\mathcal A}$, and confidence budget $\delta_r^{V}$, admission requires
\[
\widehat{\Delta A}_r\ge m_{V,r}+2W_{\mathcal A}
\sqrt{\frac{\log(2(K_r+1)/\delta_r^{V})}{2n_{V,r}}}.
\]
The frozen selection rule chooses at most one candidate passing every applicable gate as $\Delta_r^\star$ and sets $H_{r+1}=H_r\oplus\Delta_r^\star$; when the admitted set is empty, $H_{r+1}=H_r$. CineForge records each admission decision and aggregate gate statistics in an immutable audit log; replay story identifiers and artifacts do not enter CaseMemory. Across adaptive rounds, $\sum_r\delta_r^{R}\le\delta_R$, $\sum_r\delta_r^{V}\le\delta_V$, and $\delta_R+\delta_V\le\delta$. Appendix~\ref{app:experimental-protocols} specifies repeated evaluation, and Appendices~\ref{app:replay-conditions}-\ref{app:algorithms} give the full gates and deployment procedure.

\subsection{Measurable Evolution and Evaluation Boundary}
\label{sec:analysis}

Replay gates determine when CineForge-Evolve may change the persistent policy, but do not turn internal validators into a final-video quality claim. The fixed-item gate checks deterministic replay, while the following certificate extends the structural claim to fresh stories.

\begin{theorem}[Sequential fresh-replay certificate]
\label{thm:fresh-replay}
Consider adaptive rounds $r=0,\ldots,K-1$ for a fixed story distribution $\mathcal D$. Conditional on pre-admission history $\mathcal F_r$, policy $H_r$ and nonempty finite candidate set $\mathcal C_r$ are frozen before $R_r=\{\widetilde x_{r,i}\}_{i=1}^{n_{R,r}}$ is drawn conditionally i.i.d. from $\mathcal D$. Let $Y_{H_r,\Delta}(x)\in[-C,C]$, choose $\delta_r^R>0$ with $\sum_r\delta_r^R\le\delta_R$, and define
\[
G_r(\Delta)=\mathbb E[Y_{H_r,\Delta}(x)\mid\mathcal F_r],\qquad
\widehat G_r(\Delta)=\frac1{n_{R,r}}\sum_{i=1}^{n_{R,r}}Y_{H_r,\Delta}(\widetilde x_{r,i}).
\]
If CPPE accepts $\Delta_r^\star$ only when
\[
\widehat G_r(\Delta_r^\star)\ge m_r+C\sqrt{\frac{2\log(|\mathcal C_r|/\delta_r^R)}{n_{R,r}}},
\]
then, with probability at least $1-\delta_R$, every accepted round satisfies $G_r(\Delta_r^\star)\ge m_r$.
\end{theorem}

The theorem allows adaptive candidate construction across rounds but requires fresh admission stories after each candidate set is frozen. The fixed-item hard gate remains a required per-patch check; the theorem supplies an additional population statement rather than replacing it. Its cumulative structural-welfare form, definitions, and proof are given in Appendix~\ref{app:proofs}; stochastic, recurrence, and transfer results appear in Appendix~\ref{app:aux-theory}. The certificate neither establishes causal attribution nor implies rendered-video improvement. Frozen outcome evaluators therefore remain outside the complete update and selection loop, and Section~\ref{sec:datasources} establishes final-video quality $J(H)$ empirically.

\section{CineScope Benchmark}
\label{sec:cinescope}

Long-form video evaluation must assess causal development, persistent state, directorial organization, pacing, and character progression across dependent scenes, extending beyond clip-level perceptual quality. We therefore introduce \textbf{CineScope}, an evaluation framework that combines \textbf{CineScope-Data}, a 100-script suite designed to expose long-range story dependencies, with \textbf{CineScope-Metric}, a human-aligned multiscale metric for complete episodes. CineScope-Data provides evaluation inputs, while CineScope-Metric provides a frozen, output-only measure of story realization quality. We complement this suite with two public sources; detailed comparisons with existing video and narrative benchmarks are provided in Appendix~\ref{app:related-work}.

\subsection{CineScope-Data: A Diverse Collection of Long-Form Stories}
\label{sec:cinescope-data}

CineScope-Data contains 100 long-form benchmark scripts curated across diverse genres, narrative structures, character configurations, temporal spans, and scene-transition patterns. These inputs are selected to exercise cross-scene state continuity, causal development, pacing, and character progression under realistic long-video production demands. All 100 scripts are used in the reported experiments: 60 form the evolution-collection split, while the remaining 40 join 40 scripts from AnimeShooter \citep{qiu2025animeshooter} and 40 from ViStoryBench \citep{vistorybench2025} to form a 120-script non-collection pool. From each source, 10 scripts are reserved exclusively for CineScope-Metric human alignment, 15 for fresh replay admission, and 15 for final system testing, yielding mutually disjoint 30-script human-alignment, 45-script replay-admission, and 45-script system-test splits. Replay stories are revealed only after the candidate set is frozen and are used only for admission; the system-test split supplies generator comparison, controlled optimization comparison, and every round-wise unseen measurement. In every source, the complete script and its rendered episodes form the independent evaluation unit. Construction, source allocation, exact identifiers, and task-package normalization are reported in Appendix~\ref{app:cineverse}.

\subsection{CineScope-Metric: Global Evaluation for Long-Form Video}

For source narrative $x$, permitted task package $r$, and rendered episode $V$, CineScope-Metric evaluates $q=(x,r,V)$ and returns
\[
F_{\mathrm{Epi}}(q)=(\mathbf s^{S},\mathbf s^{G},\mathcal E_V),
\]
where \(\mathbf s^{S}\in[0,5]^{20}\) contains twenty directly scored criteria, \(\mathbf s^{G}\in[0,5]^4\) contains four Global aggregates, and \(\mathcal E_V\) contains temporal video evidence, eligibility decisions, and confidence. Complete-episode, 150-second, and 30-second evidence is aggregated under a criterion-by-scale contract frozen before held-out evaluation. This final-output interface is shared by all compared generators and does not expose production trajectories or CPPE state to the metric.

For criterion \(k\), the evaluator first makes a score-blind eligibility decision for every temporal window. Let $a_{k,w}\in\{0,1\}$ equal one only when the source-text evidence aligns with window $w$ and criterion $k$ is meaningfully assessable at that scope; otherwise the window is excluded for that criterion rather than scored as a failure. With window duration $d_w$, the eligible-window mean and its availability indicator are
\[
\bar s_{k,t}=\frac{\sum_{w\in\mathcal W_t}a_{k,w}d_ws_{k,w}}
{\sum_{w\in\mathcal W_t}a_{k,w}d_w},\qquad
I_{k,t}=\mathbf 1\!\left\{\sum_{w\in\mathcal W_t}a_{k,w}d_w>0\right\},
\]
for $t\in\{150,30\}$, with $\bar s_{k,t}$ omitted when $I_{k,t}=0$. The reported scores renormalize only over scopes that contain eligible evidence:
\[
s_k=\frac{\alpha_{\mathrm{epi}}s_{k,\mathrm{epi}}+
\sum_{t\in\{150,30\}}\alpha_t I_{k,t}\bar s_{k,t}}
{\alpha_{\mathrm{epi}}+\sum_{t\in\{150,30\}}\alpha_t I_{k,t}},\qquad
s_g=\sum_{k\in K_g}\beta_{gk}s_k,
\]
where $K_g$ contains the five criteria of Global dimension $g$ and $\sum_t\alpha_t=\sum_{k\in K_g}\beta_{gk}=1$. Eligibility prompts and thresholds, temporal weights $\alpha$, and within-Global weights $\beta$ are calibrated on the evaluator-development split and frozen before any held-out score is inspected. The eligibility decision is emitted before a score and cannot depend on the generator identity or score value; we report criterion-by-scale inclusion, exclusion, and abstention rates so that selective coverage remains visible.

The four dimensions target failures that independent clip averages miss:
\begin{itemize}[leftmargin=*,itemsep=1pt,topsep=2pt]
    \item \textbf{Causal State (CS)} measures whether events and state changes have coherent causes and consequences across scenes.
    \item \textbf{Directorial Orchestration (DO)} measures whether scenes, shots, camera choices, and transitions jointly serve narrative intent.
    \item \textbf{Pacing and Resource Allocation (PR)} measures whether screen time, emphasis, and editing rhythm match narrative importance.
    \item \textbf{Character Arc (CA)} measures whether identities, goals, emotions, relationships, and changes remain intelligible over the episode.
\end{itemize}
Each Global dimension aggregates five directly scored criteria. \textbf{CS:} key-event visibility; temporal-order correctness; causal-chain readability; state-change continuity; consequence follow-through and closure. \textbf{DO:} shot-function match; visual-focus clarity; plot-serving shot scale and camera movement; spatial-blocking continuity; visual emphasis and beat hierarchy. \textbf{PR:} importance-aware duration allocation; key-beat elaboration; information-density comfort; tension curve and editing rhythm; closure and breathing room. \textbf{CA:} character-identity recognizability; goal and motivation traceability; behavioral and emotional progression; relationship-dynamics progression; arc resolution and individuation.

They require \emph{global} evidence accumulated across distant scenes or the complete episode. Therefore, their twenty criteria operationalize the story-level qualities introduced in Section 1. Human raters and CineScope-Metric use the same anchored 0-5 definitions; the full rubric, aggregation equations, evidence format, confidence/abstention policy, and retry protocol appear in Appendix~\ref{app:cinescope-details}.

\paragraph{Evaluator-optimizer separation.}
As formalized in Section~\ref{sec:formulation}, CineScope-Metric is calibrated only on development data, frozen before held-out evaluation, and used exclusively for outcome reporting. It never enters CPPE's case formation, pattern induction, patch compilation, replay admission, checkpoint selection, or stopping. CPPE instead uses a separate production-trajectory reviewer with an independent prompt, model configuration, schema, and gate.

\section{Experiments}

\subsection{Experimental Setup and Evaluation Protocol}
\label{sec:datasources}

Experiments follow the three data sources and four mutually disjoint splits defined in Section~\ref{sec:cinescope-data}: 60 scripts for evolution collection, 30 for human alignment, 45 for replay admission, and 45 for final system testing. The replay-admission split is further partitioned into three disjoint round-wise sets of 15 scripts (five per source); each set is revealed only after the corresponding candidate class is frozen and supplies no case-collection or candidate-compilation signal. The system-test split never enters agent evolution and is used for generator comparison, controlled optimization comparison, and round-wise unseen evaluation.
All compared production systems use the same text and video backbones, and all CineScope-Metric evaluations use a single frozen VLM configuration; complete generation, replication, and evaluation settings are provided in Appendix~\ref{app:experimental-protocols}.
Each story-system pair is generated twice independently, and each video is scored twice by each automatic outcome evaluator; the four scores per evaluator are averaged equally into one story-level result. Round-wise checkpoints use the same globally frozen reviewer configuration and write nothing back, while subsequent evolution may still update operational thresholds and review routes. Other settings and budgets are fixed within each comparison. Appendix~\ref{app:experimental-protocols} provides the full protocol.

\subsection{CineScope-Metric Human Alignment}

This experiment calibrates CineScope-Metric and tests its alignment with expert judgments on system-unseen videos. We use the human-alignment split defined in Section~\ref{sec:datasources}, dividing each source equally between metric calibration and held-out alignment evaluation; this split is used only for metric development and held-out alignment evaluation. CineForge-Produce, MovieAgent, and AniMaker each render every script twice, yielding 180 videos. Ten trained evaluators recruited from a university score the videos with the same anchored 0–5, 20-criterion rubric used by CineScope-Metric, with at least five valid ratings per video. Human scores are averaged and aggregated into an overall score and four Global dimensions. We quantify human-metric alignment using Spearman correlation, Bradley-Terry (BT) pair agreement, and W$_1$/JS distribution similarity. The complete calibration, rating, and aggregation protocols are provided in Appendix~\ref{app:experimental-protocols}.

\begin{table}[htbp]
\centering
\scriptsize
\caption{Alignment between CineScope-Metric and held-out expert judgments at the overall and four Global-dimension levels.}
\label{tab:cinescope-validation}
\resizebox{\linewidth}{!}{%
\begin{tabular}{lcccc}
\toprule
CineScope-Metric scope & Spearman \(\rho\) [95\% CI] $\uparrow$ & BT pair agreement [95\% CI] $\uparrow$ & W$_1$ similarity [95\% CI] $\uparrow$ & JS similarity [95\% CI] $\uparrow$ \\
\midrule
Overall & 0.710 [0.550, 0.826] & 0.810 [0.698, 0.905] & 0.949 [0.926, 0.967] & 0.932 [0.823, 0.958] \\
\midrule
Causal State (CS) & 0.621 [0.463, 0.726] & 0.667 [0.524, 0.794] & 0.910 [0.883, 0.936] & 0.887 [0.797, 0.922] \\
Directorial Orchestration (DO) & 0.522 [0.342, 0.674] & 0.698 [0.571, 0.810] & 0.955 [0.921, 0.972] & 0.957 [0.859, 0.972] \\
Pacing and Resource Allocation (PR) & 0.607 [0.432, 0.742] & 0.730 [0.635, 0.825] & 0.916 [0.885, 0.941] & 0.863 [0.744, 0.901] \\
Character Arc (CA) & 0.680 [0.512, 0.806] & 0.730 [0.619, 0.841] & 0.925 [0.892, 0.947] & 0.902 [0.813, 0.937] \\
\bottomrule
\end{tabular}%
}
\end{table}

On the held-out alignment half, CineScope-Metric reaches an overall Spearman correlation of 0.710, while W$_1$ and JS similarities reach 0.949 and 0.932. Despite evaluating complete long-form episodes through multi-scale temporal evidence, CineScope-Metric maintains an 81.0\% pairwise agreement with expert judgments, comparable in magnitude to the 0.79--0.81 human-preference accuracies reported by NarrLV on its unanimous-annotator subset \citep{feng2026narrlv}. Dimension-level ranking correlations range from 0.522 to 0.680, with Character Arc showing the strongest alignment. W$_1$ and JS measure marginal distribution similarity rather than item-wise agreement. These results support CineScope-Metric as a human-aligned outcome measure; its frozen separation from CPPE also prevents evaluator feedback from driving subsequent system gains.

\subsection{Comparison of Video Generation Agents}

We compare CineForge-Produce with ViMax \citep{huang2026vimax}, AniMaker \citep{shi2025animaker}, and MovieAgent \citep{wu2025movieagent} on the system-test split under the common generation and evaluation protocol defined in Section~\ref{sec:datasources}. Further implementation and evaluation details are provided in Appendix~\ref{app:experimental-protocols}.

\begin{table}[htbp]
\centering
\scriptsize
\caption{Generator comparison on the system-test split under CineScope-Metric.}
\label{tab:generator-comparison}
\begin{tabular*}{\textwidth}{@{\extracolsep{\fill}}lccccc}
\toprule
Generator & Weighted avg. & CS & DO & PR & CA \\
\midrule
ViMax \citep{huang2026vimax} & 3.743 & 3.778 & 3.997 & 3.392 & 3.805 \\
AniMaker \citep{shi2025animaker} & 3.836 & 3.882 & 3.919 & 3.526 & 4.014 \\
MovieAgent \citep{wu2025movieagent} & 3.604 & 3.532 & 3.906 & 3.527 & 3.451 \\
\midrule
\textbf{CineForge-Produce} & \textbf{4.024} & \textbf{4.046} & \textbf{4.011} & \textbf{3.697} & \textbf{4.342} \\
\bottomrule
\end{tabular*}
\end{table}

Table~\ref{tab:generator-comparison} shows that CineForge-Produce achieves the highest CineScope-Metric weighted score (4.024) and leads all four Global dimensions, outperforming the three baselines in long-form story realization.

\FloatBarrier
\subsection{Comparison of Self-Evolution Strategies}
\label{sec:self-evolution-comparison}

To isolate the effect of the optimization method, we use CineForge-Produce as a common baseline and apply three alternative optimization mechanisms to the same production pipeline: GenMAC \citep{huang2024genmac}, VideoWeaver \citep{wei2026videoweaver}, and our CineForge-Evolve. The resulting systems are denoted CineForge-Produce + GenMAC, CineForge-Produce + VideoWeaver, and CineForge (Full), respectively. Specifically, the GenMAC and VideoWeaver variants adapt the corresponding optimization mechanisms to CineForge-Produce rather than replacing the common production pipeline with the methods' original generation pipelines. All variants start from the same CineForge-Produce checkpoint, use the evolution-collection split for method-specific optimization, and are evaluated on the system-test split under the common backbone, generation-budget, and aggregation protocol defined in Section~\ref{sec:datasources}. CineScope-Metric remains frozen and is used only for final-output evaluation.

\begin{table}[htbp]
\centering
\scriptsize
\caption{Controlled comparison of three optimization methods applied to the common CineForge-Produce baseline. All scores are measured on the system-test split under the frozen CineScope-Metric protocol.}
\label{tab:self-evolution-comparison}
\begin{tabular*}{\textwidth}{@{\extracolsep{\fill}}lccccc}
\toprule
System & Weighted avg. $\uparrow$ & CS $\uparrow$ & DO $\uparrow$ & PR $\uparrow$ & CA $\uparrow$ \\
\midrule
CineForge-Produce & 4.024 & 4.046 & 4.011 & 3.697 & 4.342 \\
CineForge-Produce + GenMAC \citep{huang2024genmac} & 4.118 & 4.171 & 4.083 & 3.791 & 4.427 \\
CineForge-Produce + VideoWeaver \citep{wei2026videoweaver} & 4.183 & 4.231 & 4.184 & 3.856 & 4.461 \\
\textbf{CineForge (Full)} & \textbf{4.380} & \textbf{4.544} & \textbf{4.372} & \textbf{3.954} & \textbf{4.650} \\
\bottomrule
\end{tabular*}
\end{table}

Because all three optimization methods share the same CineForge-Produce baseline, backbone configuration, generation budget, and final test protocol, their score differences from the first row measure the end-to-end gains of the respective optimization chains under this controlled setup rather than differences in the underlying generator. GenMAC and VideoWeaver raise the weighted score from 4.024 to 4.118 and 4.183, respectively, while CineForge (Full), which uses the Round~3 policy produced by CineForge-Evolve, reaches 4.380 and achieves the largest improvement across all four Global dimensions.

\FloatBarrier
\subsection{Self-Evolution Scaling}
\label{sec:self-evolution-experiments}

We validate how CPPE operates within CineForge-Evolve by measuring performance gains across multiple evolution rounds. The evolution-collection split is partitioned into three disjoint 20-script subsets, one for each consecutive evolution round. Round~0 denotes the unevolved CineForge-Produce checkpoint, while Rounds~1--3 denote the policies obtained after the corresponding updates. CPPE converts recurring review findings and validator evidence from each collection subset into admitted policy updates. CineScope-Metric evaluates only completed outputs at these checkpoints.

For each update round, \emph{seen} is measured by a fresh, non-writing rollout on that round's collection subset, whereas \emph{unseen} is measured on the fixed system-test split defined in Section~\ref{sec:datasources}, which is excluded from the entire evolution path. The Round~0 reference is evaluated before any update under the same read-only protocol. Every evaluation rollout is discarded and cannot update CaseMemory, CPPE, the persistent policy, checkpoint selection, or stopping. Both curves use the same globally frozen reviewer configuration, outcome evaluators, generation budgets, and two-generation/two-evaluation aggregation. Detailed split assignments, freeze points, review settings, and audit rules are provided in Appendices~\ref{app:cineverse} and~\ref{app:eval}.

\begin{figure}[!t]
\centering
\includegraphics[width=0.94\textwidth]{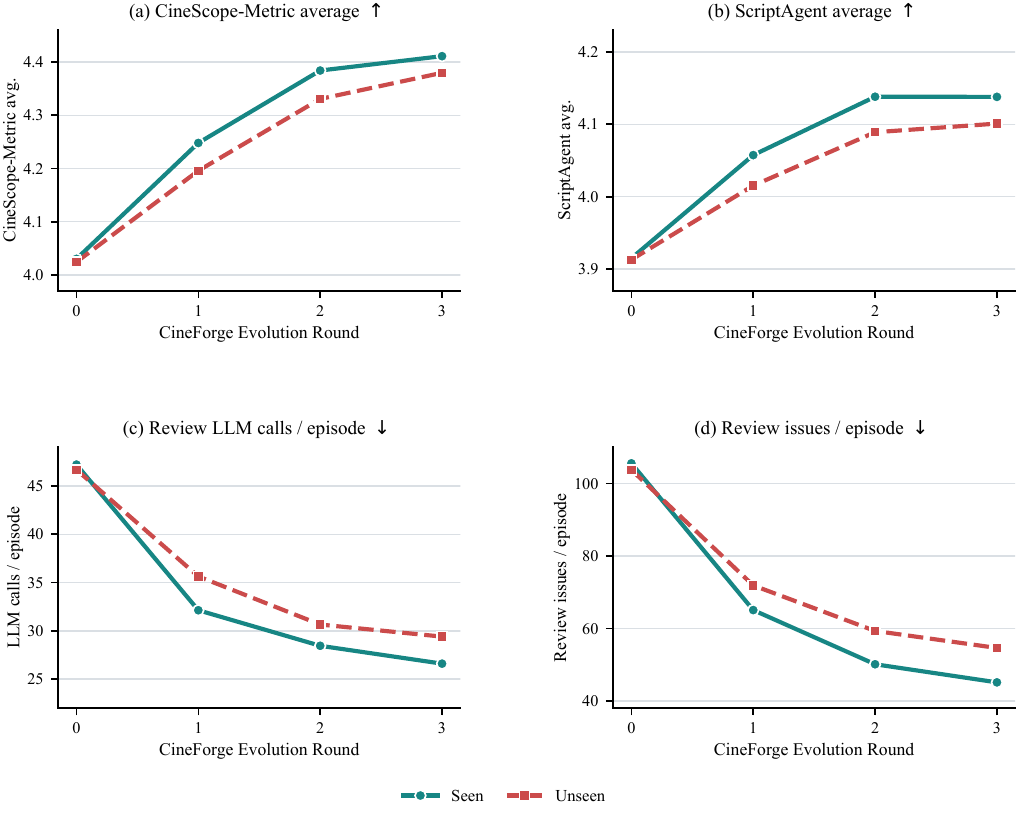}
\caption{Round-wise self-evolution from Round~0 through Round~3 for (a) CineScope-Metric, (b) ScriptAgent, (c) Review LLM calls per episode, and (d) Review issues per episode. Seen results use the corresponding round's collection subset; unseen results use the fixed system-test split, which is excluded from evolution. Round~0 is the pre-evolution reference. All checkpoint tests use the same globally frozen reviewer configuration and write nothing back.}
\label{fig:roundwise-seen-unseen}
\end{figure}

Figure~\ref{fig:roundwise-seen-unseen} shows consistent gains in outcome quality and review efficiency across Rounds~0--3. On the unseen set, CineScope-Metric rises from 4.0243 to 4.1957, 4.3306, and 4.3797 ($+0.355$), while ScriptAgent rises from 3.9130 to 4.0153, 4.0890, and 4.1009 ($+0.188$). Under the common frozen reviewer, unseen Review LLM calls fall from 46.68 to 35.63 per episode (23.7\%) and Review issues from 103.79 to 71.87 (30.8\%) by Round~1, then continue to decline. The larger seen gains in CineScope-Metric and review efficiency indicate that CPPE corrects recurring failures in each round's collection subset, while monotonic unseen improvements under both outcome evaluators demonstrate transfer beyond those data. The near-flat seen ScriptAgent score from Round~2 to Round~3 and diminishing later unseen gains suggest gradual saturation without uncontrolled overfitting. Full workload definitions and process audits are provided in Appendix~\ref{app:eval}.

\FloatBarrier

\section{Conclusion}
\label{sec:conclusion}

This work formulates long-horizon story-to-video generation as measurable cross-production self-evolution. CineForge-Produce exposes typed production trajectories while generating complete episodes, and CineForge-Evolve uses CPPE to localize unresolved process failures and convert recurring cases into bounded policy patches admitted through fresh structural replay and, when required, a stochastic admission gate. Together, they form a production-to-policy loop in which only validated experience changes future behavior. Independently, CineScope combines the CineScope-Data test suite with CineScope-Metric to evaluate episode-level causal state, directorial orchestration, pacing and resource allocation, and character arcs. Human alignment, frozen generator comparison, and round-wise seen/unseen evaluation characterize both transfer and review efficiency without conflating in-run repair, persistent policy updates, and outcome evaluation; detailed audit conditions are provided in Appendices~\ref{app:eval} and~\ref{app:replay-conditions}.

\FloatBarrier

\bibliographystyle{iclr2027_conference}
\bibliography{references}

@article{wang2024aesopagent,
  title={AesopAgent: Agent-driven Evolutionary System on Story-to-Video Production},
  author={Wang, Jiuniu and Du, Zehua and Zhao, Yuyuan and Yuan, Bo and Wang, Kexiang and Liang, Jian and Zhao, Yaxi and Lu, Yihen and Chen, Tianshi and Zhou, Qicheng and Wu, Huisi and Yu, Haoyang and Zhang, Yang and Song, Zuxuan and Xu, Jiaqi and Lin, Dahua and Wang, Limin and Qiao, Yu and Dai, Jifeng and Zhang, Wenhai},
  journal={arXiv preprint arXiv:2403.07952},
  year={2024},
  url={https://arxiv.org/abs/2403.07952}
}

@article{soni2024videoagent,
  title={VideoAgent: Self-Improving Video Generation},
  author={Soni, Achint and Venkataraman, Sreyas and Chandra, Abhranil and Fischmeister, Sebastian and Liang, Percy and Dai, Bo and Yang, Sherry},
  journal={arXiv preprint arXiv:2410.10076},
  year={2024},
  url={https://arxiv.org/abs/2410.10076}
}

@article{long2025vista,
  title={VISTA: A Test-Time Self-Improving Video Generation Agent},
  author={Long, Do Xuan and Wan, Xingchen and Nakhost, Hootan and Lee, Chen-Yu and Pfister, Tomas and Arik, Sercan {\"O}.},
  journal={arXiv preprint arXiv:2510.15831},
  year={2025},
  url={https://arxiv.org/abs/2510.15831}
}

@article{long2026a2rd,
  title={A$^2$RD: Agentic Autoregressive Diffusion for Long Video Consistency},
  author={Long, Do Xuan and Song, Yale and Kan, Min-Yen and Pfister, Tomas and Le, Long T.},
  journal={arXiv preprint arXiv:2605.06924},
  year={2026},
  url={https://arxiv.org/abs/2605.06924}
}

@article{wu2025movieagent,
  title={Automated Movie Generation via Multi-Agent CoT Planning},
  author={Wu, Weijia and Zhu, Zeyu and Shou, Mike Zheng},
  journal={arXiv preprint arXiv:2503.07314},
  year={2025},
  url={https://arxiv.org/abs/2503.07314}
}

@article{huang2024genmac,
  title={GenMAC: Compositional Text-to-Video Generation with Multi-Agent Collaboration},
  author={Huang, Kaiyi and Huang, Yukun and Ning, Xuefei and Lin, Zinan and Wang, Yu and Liu, Xihui},
  journal={arXiv preprint arXiv:2412.04440},
  year={2024},
  url={https://arxiv.org/abs/2412.04440}
}

@article{lao2026agentx,
  title={AgentX: Towards Agent-Driven Self-Iteration of Industrial Recommender Systems},
  author={Lao, Changxin and Pan, Fei and Ma, Guozhuang and Li, Han and Lin, Huihuang and Shi, Jijun and Zhao, Kangzhi and Gai, Kun and Zhou, Mo and Zhou, Qinqin and Chen, Quan and Yang, Ruochen and Bie, Shifu and Yi, Shijie and Yang, Shuang and Yang, Shuo and Li, Wenhao and Xie, Wentao and Lv, Xiao and Wang, Xuming and Wang, Yijun and Chen, Yiming and Huang, Yusheng and Wang, Zhongyuan and Zhao, Zibo and Zhuang, Zijie and others},
  journal={arXiv preprint arXiv:2606.26859},
  year={2026},
  url={https://arxiv.org/abs/2606.26859}
}

@article{han2025mapgd,
  title={MAPGD: Multi-Agent Prompt Gradient Descent for Collaborative Prompt Optimization},
  author={Han, Yichen and Liu, Bojun and Zhou, Zhengpeng and Zhou, Guanyu and Zhang, Zeng and Yang, Yang and Wang, Wenli and Shi, Isaac N. and Yunyan and He, Lewei and Shi, Tianyu},
  journal={arXiv preprint arXiv:2509.11361},
  year={2025},
  url={https://arxiv.org/abs/2509.11361}
}

@article{madaan2023selfrefine,
  title={Self-Refine: Iterative Refinement with Self-Feedback},
  author={Madaan, Aman and Tandon, Niket and Gupta, Prakhar and Hallinan, Skyler and Gao, Luyu and Wiegreffe, Sarah and Alon, Uri and Dziri, Nouha and Prabhumoye, Shrimai and Yang, Yiming and others},
  journal={arXiv preprint arXiv:2303.17651},
  year={2023},
  url={https://arxiv.org/abs/2303.17651}
}

@inproceedings{shinn2024reflexion,
  title={Reflexion: Language Agents with Verbal Reinforcement Learning},
  author={Shinn, Noah and Cassano, Federico and Gopinath, Ashwin and Narasimhan, Karthik and Yao, Shunyu},
  booktitle={Advances in Neural Information Processing Systems},
  year={2024}
}

@article{wang2023voyager,
  title={Voyager: An Open-Ended Embodied Agent with Large Language Models},
  author={Wang, Guanzhi and Xie, Yuqi and Jiang, Yunfan and Mandlekar, Ajay and Xiao, Chaowei and Zhu, Yuke and Fan, Linxi and Anandkumar, Anima},
  journal={arXiv preprint arXiv:2305.16291},
  year={2023},
  url={https://arxiv.org/abs/2305.16291}
}

@inproceedings{zhou2023ape,
  title={Large Language Models Are Human-Level Prompt Engineers},
  author={Zhou, Yongchao and Muresanu, Andrei Ioan and Han, Ziwen and Paster, Keiran and Pitis, Silviu and Chan, Harris and Ba, Jimmy},
  booktitle={International Conference on Learning Representations},
  year={2023},
  url={https://arxiv.org/abs/2211.01910}
}

@article{pryzant2023automatic,
  title={Automatic Prompt Optimization with ``Gradient Descent'' and Beam Search},
  author={Pryzant, Reid and Iter, Dan and Li, Jerry and Lee, Yin Tat and Zhu, Chenguang and Zeng, Michael},
  journal={arXiv preprint arXiv:2305.03495},
  year={2023},
  url={https://arxiv.org/abs/2305.03495}
}

@inproceedings{yang2024opro,
  title={Large Language Models as Optimizers},
  author={Yang, Chengrun and Wang, Xuezhi and Lu, Yifeng and Liu, Hanxiao and Le, Quoc V. and Zhou, Denny and Chen, Xinyun},
  booktitle={International Conference on Learning Representations},
  year={2024},
  url={https://arxiv.org/abs/2309.03409}
}

@article{yuksekgonul2024textgrad,
  title={TextGrad: Automatic ``Differentiation'' via Text},
  author={Yuksekgonul, Mert and Bianchi, Federico and Boen, Joseph and Liu, Sheng and Huang, Zhi and Guestrin, Carlos and Zou, James},
  journal={arXiv preprint arXiv:2406.07496},
  year={2024},
  url={https://arxiv.org/abs/2406.07496}
}

@article{khattab2023dspy,
  title={DSPy: Compiling Declarative Language Model Calls into Self-Improving Pipelines},
  author={Khattab, Omar and Singhvi, Arnav and Maheshwari, Paridhi and Zhang, Zhiyuan and Santhanam, Keshav and Vardhamanan, Sri and Haq, Saiful and Sharma, Ashutosh and Joshi, Thomas T. and Moazam, Hanna and others},
  journal={arXiv preprint arXiv:2310.03714},
  year={2023},
  url={https://arxiv.org/abs/2310.03714}
}

@inproceedings{kakade2002approximately,
  title={Approximately Optimal Approximate Reinforcement Learning},
  author={Kakade, Sham and Langford, John},
  booktitle={International Conference on Machine Learning},
  pages={267--274},
  year={2002}
}

@inproceedings{pirotta2013safe,
  title={Safe Policy Iteration},
  author={Pirotta, Matteo and Restelli, Marcello and Pecorino, Alessio and Calandriello, Daniele},
  booktitle={International Conference on Machine Learning},
  pages={307--315},
  year={2013}
}

@inproceedings{thomas2015high,
  title={High Confidence Policy Improvement},
  author={Thomas, Philip and Theocharous, Georgios and Ghavamzadeh, Mohammad},
  booktitle={International Conference on Machine Learning},
  pages={2380--2388},
  year={2015}
}

@book{boucheron2013concentration,
  title={Concentration Inequalities: A Nonasymptotic Theory of Independence},
  author={Boucheron, St{\'e}phane and Lugosi, G{\'a}bor and Massart, Pascal},
  publisher={Oxford University Press},
  year={2013}
}

@article{hoeffding1963probability,
  title={Probability Inequalities for Sums of Bounded Random Variables},
  author={Hoeffding, Wassily},
  journal={Journal of the American Statistical Association},
  volume={58},
  number={301},
  pages={13--30},
  year={1963}
}

@article{legoues2019automated,
  title={Automated Program Repair},
  author={Le Goues, Claire and Pradel, Michael and Roychoudhury, Abhik},
  journal={Communications of the ACM},
  volume={62},
  number={12},
  pages={56--65},
  year={2019}
}

@inproceedings{huang2020movienet,
  title={MovieNet: A Holistic Dataset for Movie Understanding},
  author={Huang, Qingqiu and Xiong, Yu and Rao, Anyi and Wang, Jiaze and Lin, Dahua},
  booktitle={European Conference on Computer Vision},
  pages={709--727},
  year={2020}
}

@inproceedings{li2019storygan,
  title={StoryGAN: A Sequential Conditional GAN for Story Visualization},
  author={Li, Yitong and Gan, Zhe and Shen, Yelong and Liu, Jingjing and Cheng, Yu and Wu, Yuexin and Carin, Lawrence and Carlson, David and Gao, Jianfeng},
  booktitle={Proceedings of the IEEE/CVF Conference on Computer Vision and Pattern Recognition},
  pages={6329--6338},
  year={2019},
  url={https://arxiv.org/abs/1812.02784}
}

@inproceedings{rahman2023makeastory,
  title={Make-A-Story: Visual Memory Conditioned Consistent Story Generation},
  author={Rahman, Tanzila and Lee, Hsin-Ying and Ren, Jian and Tulyakov, Sergey and Mahajan, Shweta and Sigal, Leonid},
  booktitle={Proceedings of the IEEE/CVF Conference on Computer Vision and Pattern Recognition},
  year={2023},
  url={https://arxiv.org/abs/2211.13319}
}

@article{lin2023videodirectorgpt,
  title={VideoDirectorGPT: Consistent Multi-scene Video Generation via LLM-Guided Planning},
  author={Lin, Han and Cho, Jaemin and Zala, Abhay and Bansal, Mohit},
  journal={arXiv preprint arXiv:2309.15091},
  year={2023},
  url={https://arxiv.org/abs/2309.15091}
}

@article{pan2026rho,
  title={Evolving Agents in the Dark: Retrospective Harness Optimization via Self-Preference},
  author={Pan, Wenbo and Liu, Shujie and Lin, Chin-Yew and Zeng, Jingying and Tang, Xianfeng and Zhou, Xiangyang and Lu, Yan and Jia, Xiaohua},
  journal={arXiv preprint arXiv:2606.05922},
  year={2026},
  url={https://arxiv.org/abs/2606.05922}
}

@article{pan2026nlah,
  title={Natural-Language Agent Harnesses},
  author={Pan, Linyue and Zou, Lexiao and Guo, Shuo and Ni, Jingchen and Zheng, Hai-Tao},
  journal={arXiv preprint arXiv:2603.25723},
  year={2026},
  url={https://arxiv.org/abs/2603.25723}
}

@article{lin2026ahe,
  title={Agentic Harness Engineering: Observability-Driven Automatic Evolution of Coding-Agent Harnesses},
  author={Lin, Jiahang and Liu, Shichun and Pan, Chengjun and Lin, Lizhi and Dou, Shihan and Xi, Zhiheng and Huang, Xuanjing and Yan, Hang and Han, Zhenhua and Gui, Tao and Jiang, Yu-Gang},
  journal={arXiv preprint arXiv:2604.25850},
  year={2026},
  url={https://arxiv.org/abs/2604.25850}
}

@article{chen2026harnessfix,
  title={From Failed Trajectories to Reliable LLM Agents: Diagnosing and Repairing Harness Flaws},
  author={Chen, Mengzhuo and Wang, Junjie and Liu, Zhe and Wang, Yawen and Zheng, Haiming and Wang, Qing},
  journal={arXiv preprint arXiv:2606.06324},
  year={2026},
  url={https://arxiv.org/abs/2606.06324}
}

@inproceedings{huang2024vbench,
  title={VBench: Comprehensive Benchmark Suite for Video Generative Models},
  author={Huang, Ziqi and He, Yinan and Yu, Jiashuo and Zhang, Fan and Si, Chenyang and Jiang, Yuming and Zhang, Yuanhan and Wu, Tianxing and Jin, Qingyang and Chanpaisit, Nattapol and others},
  booktitle={Proceedings of the IEEE/CVF Conference on Computer Vision and Pattern Recognition},
  year={2024},
  url={https://arxiv.org/abs/2311.17982}
}

@article{directorbench2026,
  title={DirectorBench: Diagnosing Long-Form Video Generation with Personalized Multi-Agent Evaluation},
  author={Chen, Jiamin and Chen, Qianben and Zhang, Jiawen and Wu, Yidi and Li, Yuchen and Zhang, Xiaokun and Zhou, Wangchunshu and Ma, Chen},
  journal={arXiv preprint arXiv:2605.30090},
  year={2026},
  url={https://arxiv.org/abs/2605.30090}
}

@inproceedings{msvbench2026,
  title={MSVBench: Towards Human-Level Evaluation of Multi-Shot Video Generation},
  author={Shi, Haoyuan and Li, Yunxin and Deng, Nanhao and Xu, Zhenran and Chen, Xinyu and Wang, Longyue and Hu, Baotian and Zhang, Min},
  booktitle={Findings of the Association for Computational Linguistics: ACL 2026},
  pages={24034--24058},
  publisher={Association for Computational Linguistics},
  year={2026},
  doi={10.18653/v1/2026.findings-acl.1203},
  url={https://aclanthology.org/2026.findings-acl.1203/}
}

@article{msavbench2026,
  title={MSAVBench: Towards Comprehensive and Reliable Evaluation of Multi-Shot Audio-Video Generation},
  author={Wei, Yujie and Han, Yujin and Chen, Zhekai and Li, Yongming and Jiang, Kaixun and others},
  journal={arXiv preprint arXiv:2605.20183},
  year={2026},
  url={https://arxiv.org/abs/2605.20183}
}

@article{liu2026longavcompass,
  title={{LongAV-Compass}: Towards Unified Evaluation of Minute-Scale Audio-Visual Generation Across {T2AV}, {I2AV}, and {V2AV}},
  author={Liu, Tengfei and Shi, Yang and Zhu, Xuanyu and Tang, Jiafu and Yang, Liu and Wang, Qixun and Zhang, Zhuoran and Tang, Yuqi and Wang, Fengxiang and Dong, Yuhao and Chen, Xinlong and Li, Bozhou and Zeng, Bohan and Ding, Yue and Zhang, Xiaohan and Chen, Jialu and Wang, Haotian and Zhang, Yuanxing and Wan, Pengfei and Wang, Leye},
  journal={arXiv preprint arXiv:2605.26244},
  year={2026},
  url={https://arxiv.org/abs/2605.26244}
}

@inproceedings{zheng2026locot2vbench,
  title={{LoCoT2V-Bench}: Benchmarking Long-Form and Complex Text-to-Video Generation},
  author={Zheng, Xiangqing and Wu, Chengyue and Chen, Kehai and Zhang, Min},
  booktitle={Proceedings of the 43rd International Conference on Machine Learning},
  year={2026},
  url={https://arxiv.org/abs/2510.26412}
}

@article{liu2026narrastream,
  title={Advancing Narrative Long Video Generation via Training-Free Identity-Aware Memory},
  author={Liu, Jinzhuo and Zhang, Jiangning and Jiang, Wencan and Wang, Yabiao and Liang, Dingkang and Xue, Zhucun and Yi, Ran and Liu, Yong},
  journal={arXiv preprint arXiv:2605.18733},
  year={2026},
  url={https://arxiv.org/abs/2605.18733}
}

@article{he2026entitybench,
  title={{EntityBench}: Towards Entity-Consistent Long-Range Multi-Shot Video Generation},
  author={He, Ruozhen and Wei, Meng and Yang, Ziyan and Ordonez, Vicente},
  journal={arXiv preprint arXiv:2605.15199},
  year={2026},
  url={https://arxiv.org/abs/2605.15199}
}

@inproceedings{feng2026narrlv,
  title={{NarrLV}: Towards a Comprehensive Narrative-Centric Evaluation for Long Video Generation},
  author={Feng, Xiaokun and Yu, Haiming and Wu, Meiqi and Hu, Shiyu and Chen, Jintao and Zhu, Chen and Wu, Jiahong and Chu, Xiangxiang and Huang, Kaiqi},
  booktitle={International Conference on Learning Representations},
  year={2026},
  url={https://openreview.net/forum?id=Qh3CQBTB1g}
}

@inproceedings{matsuda2026slvmeval,
  title={{SLVMEval}: Synthetic Meta Evaluation Benchmark for Text-to-Long Video Generation},
  author={Matsuda, Ryosuke and Kudo, Keito and Yoshida, Haruto and Shimizu, Nobuyuki and Suzuki, Jun},
  booktitle={Proceedings of the IEEE/CVF Conference on Computer Vision and Pattern Recognition},
  pages={7784--7794},
  year={2026},
  url={https://openaccess.thecvf.com/content/CVPR2026/papers/Matsuda_SLVMEval_Synthetic_Meta_Evaluation_Benchmark_for_Text-to-Long_Video_Generation_CVPR_2026_paper.pdf}
}

@article{wang2026filmbench,
  title={{FilmBench}: A Film-Grade Benchmark for Cinematic Video Generation},
  author={Wang, Shengyi and Li, Niantong and Hu, Guangzheng and Qi, Hong and Ding, Fei and Qiao, Weixu and Wang, Jinlin and Lv, Xiaotong and Han, Peng and Li, Zimeng and Ding, Fanshu and Wang, Yushu and Wu, Han and Chen, Jingjing and Wang, Chongxiao and Wu, Yanhao and Huang, Chenglong and Zhu, Xiaoqian and Tian, Jie and Li, Hua and Fan, Jingjing and Tang, Mingshuang and Li, Zhong and Qiang, Hengxia and Chen, Weibin and Zhen, Jinyang and Zhao, Bing and Qu, Lin and Li, Jing and Wei, Hu},
  journal={arXiv preprint arXiv:2607.24241},
  year={2026},
  url={https://arxiv.org/abs/2607.24241}
}

@article{vistorybench2025,
  title={ViStoryBench: Comprehensive Benchmark Suite for Story Visualization},
  author={Zhuang, Cailin and Huang, Ailin and Hu, Yaoqi and Wu, Jingwei and Cheng, Wei and Liao, Jiaqi and Wang, Hongyuan and Liao, Xinyao and Cai, Weiwei and Xu, Hengyuan and Zhang, Xuanyang and Zeng, Xianfang and Huang, Zhewei and Yu, Gang and Zhang, Chi},
  journal={Proceedings of the IEEE/CVF Conference on Computer Vision and Pattern Recognition},
  year={2026},
  url={https://arxiv.org/abs/2505.24862}
}

@article{gong2023talecrafter,
  title={TaleCrafter: Interactive Story Visualization with Multiple Characters},
  author={Gong, Yuan and Pang, Youxin and Cun, Xiaodong and Xia, Menghan and He, Yingqing and Chen, Haoxin and Wang, Longyue and Zhang, Yong and Wang, Xintao and Shan, Ying and Yang, Yujiu},
  journal={ACM Transactions on Graphics},
  volume={42},
  number={6},
  year={2023},
  doi={10.1145/3610548.3618184},
  url={https://arxiv.org/abs/2305.18247}
}

@inproceedings{zhou2024storydiffusion,
  title={StoryDiffusion: Consistent Self-Attention for Long-Range Image and Video Generation},
  author={Zhou, Yupeng and Zhou, Daquan and Cheng, Ming-Ming and Feng, Jiashi and Hou, Qibin},
  booktitle={Advances in Neural Information Processing Systems},
  year={2024},
  url={https://arxiv.org/abs/2405.01434}
}

@article{yang2024seedstory,
  title={SEED-Story: Multimodal Long Story Generation with Large Language Model},
  author={Yang, Shuai and Ge, Yuying and Li, Yang and Chen, Yukang and Ge, Yixiao and Shan, Ying and Chen, Yingcong},
  journal={arXiv preprint arXiv:2407.08683},
  year={2024},
  url={https://arxiv.org/abs/2407.08683}
}

@article{hu2024adas,
  title={Automated Design of Agentic Systems},
  author={Hu, Shengran and Lu, Cong and Clune, Jeff},
  journal={arXiv preprint arXiv:2408.08435},
  year={2024},
  url={https://arxiv.org/abs/2408.08435}
}

@article{zhang2024aflow,
  title={AFlow: Automating Agentic Workflow Generation},
  author={Zhang, Jiayi and Xiang, Jinyu and Yu, Zhaoyang and Teng, Fengwei and Chen, Xionghui and Chen, Jiaqi and Zhuge, Mingchen and Cheng, Xin and Hong, Sirui and Wang, Jinlin and Zheng, Bingnan and Liu, Bang and Luo, Yuyu and Wu, Chenglin},
  journal={arXiv preprint arXiv:2410.10762},
  year={2024},
  url={https://arxiv.org/abs/2410.10762}
}

@article{yang2026skillopt,
  title={SkillOpt: Executive Strategy for Self-Evolving Agent Skills},
  author={Yang, Yifan and Gong, Ziyang and Huang, Weiquan and Yang, Qihao and Zhou, Ziwei and Huang, Zisu and Li, Yan and Gao, Xuemei and Dai, Qi and Liu, Bei and Qiu, Kai and Yang, Yuqing and Chen, Dongdong and Yang, Xue and Luo, Chong},
  journal={arXiv preprint arXiv:2605.23904},
  year={2026},
  url={https://arxiv.org/abs/2605.23904}
}

@inproceedings{kim2024fifo,
  title={{FIFO-Diffusion}: Generating Infinite Videos from Text without Training},
  author={Kim, Jihwan and Kang, Junoh and Choi, Jinyoung and Han, Bohyung},
  booktitle={Advances in Neural Information Processing Systems},
  volume={37},
  year={2024},
  doi={10.52202/079017-2853},
  url={https://arxiv.org/abs/2405.11473}
}

@inproceedings{lu2024freelong,
  title={FreeLong: Training-Free Long Video Generation with SpectralBlend Temporal Attention},
  author={Lu, Yu and Liang, Yuanzhi and Zhu, Linchao and Yang, Yi},
  booktitle={Advances in Neural Information Processing Systems},
  volume={37},
  year={2024},
  doi={10.52202/079017-4177},
  url={https://proceedings.neurips.cc/paper_files/paper/2024/hash/ed67dff7cb96e7e86c4d91c0d5db49bb-Abstract.html}
}

@inproceedings{guo2025longcontext,
  title={Long Context Tuning for Video Generation},
  author={Guo, Yuwei and Yang, Ceyuan and Yang, Ziyan and Ma, Zhibei and Lin, Zhijie and Yang, Zhenheng and Lin, Dahua and Jiang, Lu},
  booktitle={Proceedings of the IEEE/CVF International Conference on Computer Vision},
  pages={17281--17291},
  year={2025},
  url={https://openaccess.thecvf.com/content/ICCV2025/html/Guo_Long_Context_Tuning_for_Video_Generation_ICCV_2025_paper.html}
}

@inproceedings{wu2025moviebench,
  title={MovieBench: A Hierarchical Movie Level Dataset for Long Video Generation},
  author={Wu, Weijia and Liu, Mingyu and Zhu, Zeyu and Xia, Xi and Feng, Haoen and Wang, Wen and Lin, Kevin Qinghong and Shen, Chunhua and Shou, Mike Zheng},
  booktitle={Proceedings of the IEEE/CVF Conference on Computer Vision and Pattern Recognition},
  pages={28984--28994},
  year={2025},
  url={https://openaccess.thecvf.com/content/CVPR2025/html/Wu_MovieBench_A_Hierarchical_Movie_Level_Dataset_for_Long_Video_Generation_CVPR_2025_paper.html}
}

@inproceedings{kara2025shotadapter,
  title={ShotAdapter: Text-to-Multi-Shot Video Generation with Diffusion Models},
  author={Kara, Ozgur and Singh, Krishna Kumar and Liu, Feng and Ceylan, Duygu and Rehg, James M. and Hinz, Tobias},
  booktitle={Proceedings of the IEEE/CVF Conference on Computer Vision and Pattern Recognition},
  pages={28405--28415},
  year={2025},
  url={https://openaccess.thecvf.com/content/CVPR2025/html/Kara_ShotAdapter_Text-to-Multi-Shot_Video_Generation_with_Diffusion_Models_CVPR_2025_paper.html}
}

@inproceedings{hong2024metagpt,
  title={MetaGPT: Meta Programming for A Multi-Agent Collaborative Framework},
  author={Hong, Sirui and Zhuge, Mingchen and Chen, Jonathan and Zheng, Xiawu and Cheng, Yuheng and Wang, Jinlin and Zhang, Ceyao and Wang, Zili and Yau, Steven and Lin, Zijuan and Zhou, Liyang and Ran, Chenyu and Xiao, Lingfeng and Wu, Chenglin and Schmidhuber, J{\"u}rgen},
  booktitle={International Conference on Learning Representations},
  year={2024},
  url={https://openreview.net/forum?id=VtmBAGCN7o}
}

@inproceedings{shrimal2024marco,
  title={{MARCO}: Multi-Agent Real-time Chat Orchestration},
  author={Shrimal, Anubhav and Kanagaraj, Stanley and Biswas, Kriti and Raghuraman, Swarnalatha and Nediyanchath, Anish and Zhang, Yi and Yenigalla, Promod},
  booktitle={Proceedings of the 2024 Conference on Empirical Methods in Natural Language Processing: Industry Track},
  pages={1381--1392},
  year={2024},
  doi={10.18653/v1/2024.emnlp-industry.102},
  url={https://aclanthology.org/2024.emnlp-industry.102/}
}

@inproceedings{zhao2024expel,
  title={ExpeL: LLM Agents Are Experiential Learners},
  author={Zhao, Andrew and Huang, Daniel and Xu, Quentin and Lin, Matthieu and Liu, Yong-Jin and Huang, Gao},
  booktitle={Proceedings of the AAAI Conference on Artificial Intelligence},
  volume={38},
  pages={19632--19642},
  year={2024},
  doi={10.1609/aaai.v38i17.29936},
  url={https://ojs.aaai.org/index.php/AAAI/article/view/29936}
}

@inproceedings{zhang2026ace,
  title={Agentic Context Engineering: Evolving Contexts for Self-Improving Language Models},
  author={Zhang, Qizheng and Hu, Changran and Upasani, Shubhangi and Ma, Boyuan and Hong, Fenglu and Kamanuru, Vamsidhar and Rainton, Jay and Wu, Chen and Ji, Mengmeng and Li, Hanchen and Thakker, Urmish and Zou, James and Olukotun, Kunle},
  booktitle={International Conference on Learning Representations},
  year={2026},
  url={https://iclr.cc/virtual/2026/poster/10008343}
}

@inproceedings{agrawal2026gepa,
  title={{GEPA}: Reflective Prompt Evolution Can Outperform Reinforcement Learning},
  author={Agrawal, Lakshya A. and Tan, Shangyin and Soylu, Dilara and Ziems, Noah and Khare, Rishi and Opsahl-Ong, Krista and Singhvi, Arnav and Shandilya, Herumb and Ryan, Michael J. and Jiang, Meng and Potts, Christopher and Sen, Koushik and Dimakis, Alexandros G. and Stoica, Ion and Klein, Dan and Zaharia, Matei and Khattab, Omar},
  booktitle={International Conference on Learning Representations},
  year={2026},
  url={https://iclr.cc/virtual/2026/poster/10009493}
}

@inproceedings{opsahlong2024mipro,
  title={{MIPRO}: Optimizing Instructions and Demonstrations for Multi-Stage {LM} Programs},
  author={Opsahl-Ong, Krista and Ryan, Michael J. and Purtell, Josh and Broman, David and Potts, Christopher and Zaharia, Matei and Khattab, Omar},
  booktitle={Proceedings of the 2024 Conference on Empirical Methods in Natural Language Processing},
  pages={9340--9366},
  year={2024},
  doi={10.18653/v1/2024.emnlp-main.525},
  url={https://aclanthology.org/2024.emnlp-main.525/}
}

@article{nie2026tthe,
  title={{TTHE}: Test-Time Harness Evolution},
  author={Nie, Jun and Zhang, Yonggang and Song, Jun and Cai, Qianshu and Yu, Dahai and Guo, Yike and Tian, Xinmei and Han, Bo},
  journal={arXiv preprint arXiv:2607.08124},
  year={2026},
  url={https://arxiv.org/abs/2607.08124}
}

@article{lou2026autoharness,
  title={AutoHarness: Improving LLM Agents by Automatically Synthesizing a Code Harness},
  author={Lou, Xinghua and L{\'a}zaro-Gredilla, Miguel and Dedieu, Antoine and Wendelken, Carter and Lehrach, Wolfgang and Murphy, Kevin P.},
  journal={arXiv preprint arXiv:2603.03329},
  year={2026},
  url={https://arxiv.org/abs/2603.03329}
}

@article{che2026demoevolve,
  title={DemoEvolve: Overcoming Sparse Feedback in Agentic Harness Evolution with Demonstrations},
  author={Che, Lirong and Yang, Yuzhe and Lin, Peiwen and Wang, Chuang and Wang, Xueqian and Su, Jian},
  journal={arXiv preprint arXiv:2605.24539},
  year={2026},
  url={https://arxiv.org/abs/2605.24539}
}

@article{lin2026harnessbenefit,
  title={Harness Updating Is Not Harness Benefit: Disentangling Evolution Capabilities in Self-Evolving LLM Agents},
  author={Lin, Minhua and Wu, Juncheng and Wang, Zijun and Shi, Zhan and Sang, Yisi and He, Bing and Liu, Zewen and Wei, Tianxin and Wu, Zongyu and Zhang, Zhiwei and Wang, Dakuo and Zhang, Xiang and Dumoulin, Benoit and Xie, Cihang and Zhou, Yuyin and Wang, Suhang and Lu, Hanqing},
  journal={arXiv preprint arXiv:2605.30621},
  year={2026},
  url={https://arxiv.org/abs/2605.30621}
}

@article{wang2026rethinkingharnesseval,
  title={Rethinking the Evaluation of Harness Evolution for Agents},
  author={Wang, Yike and Zhu, Huaisheng and Hu, Zhengyu and Yuan, Yige and Chen, Zhengyu and Senthil, Shakti and Hajishirzi, Hannaneh and Tsvetkov, Yulia and Dasigi, Pradeep and Xiao, Teng},
  journal={arXiv preprint arXiv:2607.12227},
  year={2026},
  url={https://arxiv.org/abs/2607.12227}
}

@inproceedings{laroche2019spibb,
  title={Safe Policy Improvement with Baseline Bootstrapping},
  author={Laroche, Romain and Trichelair, Paul and Combes, Remi Tachet Des},
  booktitle={Proceedings of the 36th International Conference on Machine Learning},
  series={Proceedings of Machine Learning Research},
  volume={97},
  pages={3652--3661},
  year={2019},
  url={https://proceedings.mlr.press/v97/laroche19a.html}
}

@inproceedings{bugliarello2023storybench,
  title={StoryBench: A Multifaceted Benchmark for Continuous Story Visualization},
  author={Bugliarello, Emanuele and Moraldo, Hernan and Villegas, Ruben and Babaeizadeh, Mohammad and Saffar, Mohammad Taghi and Zhang, Han and Erhan, Dumitru and Ferrari, Vittorio and Kindermans, Pieter-Jan and Voigtlaender, Paul},
  booktitle={Advances in Neural Information Processing Systems},
  volume={36},
  pages={78095--78125},
  year={2023},
  url={https://proceedings.neurips.cc/paper_files/paper/2023/hash/f63f5fbed1a4ef08c857c5f377b5d33a-Abstract-Datasets_and_Benchmarks.html}
}

@inproceedings{wang2025storyeval,
  title={Is Your World Simulator a Good Story Presenter? A Consecutive Events-Based Benchmark for Future Long Video Generation},
  author={Wang, Yiping and He, Xuehai and Wang, Kuan and Ma, Luyao and Yang, Jianwei and Wang, Shuohang and Du, Simon Shaolei and Shen, Yelong},
  booktitle={Proceedings of the IEEE/CVF Conference on Computer Vision and Pattern Recognition},
  pages={13629--13638},
  year={2025},
  url={https://openaccess.thecvf.com/content/CVPR2025/html/Wang_Is_Your_World_Simulator_a_Good_Story_Presenter_A_Consecutive_CVPR_2025_paper.html}
}

@inproceedings{liu2024evalcrafter,
  title={EvalCrafter: Benchmarking and Evaluating Large Video Generation Models},
  author={Liu, Yaofang and Cun, Xiaodong and Liu, Xuebo and Wang, Xintao and Zhang, Yong and Chen, Haoxin and Liu, Yang and Zeng, Tieyong and Chan, Raymond and Shan, Ying},
  booktitle={Proceedings of the IEEE/CVF Conference on Computer Vision and Pattern Recognition},
  pages={22139--22149},
  year={2024},
  url={https://openaccess.thecvf.com/content/CVPR2024/html/Liu_EvalCrafter_Benchmarking_and_Evaluating_Large_Video_Generation_Models_CVPR_2024_paper.html}
}

@inproceedings{tapaswi2015book2movie,
  title={Book2Movie: Aligning Video Scenes with Book Chapters},
  author={Tapaswi, Makarand and B{\"a}uml, Martin and Stiefelhagen, Rainer},
  booktitle={Proceedings of the IEEE Conference on Computer Vision and Pattern Recognition},
  pages={1827--1835},
  year={2015},
  doi={10.1109/CVPR.2015.7298792},
  url={https://openaccess.thecvf.com/content_cvpr_2015/html/Tapaswi_Book2Movie_Aligning_Video_2015_CVPR_paper.html}
}

@article{qiu2025animeshooter,
  title={AnimeShooter: A Multi-Shot Animation Dataset for Reference-Guided Video Generation},
  author={Qiu, Lu and Li, Yizhuo and Ge, Yuying and Ge, Yixiao and Shan, Ying and Liu, Xihui},
  journal={arXiv preprint arXiv:2506.03126},
  year={2025},
  url={https://arxiv.org/abs/2506.03126}
}

@article{wei2026videoweaver,
  title={VideoWeaver: Evaluating and Evolving Skills for Agentic Long Video Generation},
  author={Wei, Jianhui and Tan, Jie and Zhu, Hengchuan and Zhang, Xiaotian and Zhang, Yan and Chen, Ziyi and Zhang, Daoan and Xu, Wei and Liu, Zuozhu},
  journal={arXiv preprint arXiv:2606.08091},
  year={2026},
  url={https://arxiv.org/abs/2606.08091}
}

@article{mu2026scriptagent,
  title={The Script Is All You Need: An Agentic Framework for Long-Horizon Dialogue-to-Cinematic Video Generation},
  author={Mu, Chenyu and He, Xin and Yang, Qu and Chen, Wanshun and Yao, Jiadi and Liu, Huang and Yi, Zihao and Zhao, Bo and Chen, Xingyu and Ma, Ruotian and Ye, Fanghua and Yang, Erkun and Deng, Cheng and Tu, Zhaopeng and Li, Xiaolong and Linus},
  journal={arXiv preprint arXiv:2601.17737},
  year={2026},
  url={https://arxiv.org/abs/2601.17737}
}

@article{huang2024vbenchpp,
  title={{VBench++}: Comprehensive and Versatile Benchmark Suite for Video Generative Models},
  author={Huang, Ziqi and Zhang, Fan and Xu, Xiaojie and He, Yinan and Yu, Jiashuo and Dong, Ziyue and Ma, Qianli and Chanpaisit, Nattapol and Si, Chenyang and Jiang, Yuming and others},
  journal={arXiv preprint arXiv:2411.13503},
  year={2024},
  url={https://arxiv.org/abs/2411.13503}
}

@article{huang2026vimax,
  title={{ViMax}: Agentic Video Generation},
  author={Huang, Lingxuan and He, Sizhe and Zhou, Hengji and Nie, Liqiang and Xia, Lianghao and Huang, Chao},
  journal={arXiv preprint arXiv:2606.07649},
  year={2026},
  url={https://arxiv.org/abs/2606.07649}
}

@article{shi2025animaker,
  title={{AniMaker}: Multi-Agent Animated Storytelling with {MCTS}-Driven Clip Generation},
  author={Shi, Haoyuan and Li, Yunxin and Chen, Xinyu and Wang, Longyue and Hu, Baotian and Zhang, Min},
  journal={arXiv preprint arXiv:2506.10540},
  year={2025},
  url={https://arxiv.org/abs/2506.10540}
}

\appendix

\section{Detailed Related Work}
\label{app:related-work}

\subsection{Long-Horizon Story and Video Generation}

Recent video-generation systems increasingly use language models or agents to decompose a high-level description into scenes, shots, camera plans, or multi-agent production roles. Methods such as VideoDirectorGPT, MovieAgent, GenMAC, ScriptAgent, StoryDiffusion, and SEED-Story show the value of planning, memory, or consistency mechanisms for multi-scene visual generation \citep{lin2023videodirectorgpt,wu2025movieagent,huang2024genmac,mu2026scriptagent,zhou2024storydiffusion,yang2024seedstory,hong2024metagpt,shrimal2024marco}. These systems motivate our setting, but they optimize a different object. Their main goal is to produce better plans or videos for the current input; our focus is on making repeated production failures measurable and using them to update the persistent rule layer of the workflow. The distinction matters for long-horizon production: a system that repairs one failed clip is useful, but a system that prevents the same structural failure from recurring across later stories changes the behavior of the production process itself.

Earlier story-visualization systems established sequential generation, visual memory, and multi-character consistency as central requirements for narrative visual synthesis \citep{li2019storygan,rahman2023makeastory,gong2023talecrafter}. Long-video methods further extend temporal context or introduce shot-aware adaptation, but do not by themselves define how recurring cross-stage production failures should update a persistent workflow \citep{kim2024fifo,lu2024freelong,guo2025longcontext,kara2025shotadapter}.

\subsection{Long-Video and Narrative Evaluation}

Rendered-output suites such as VBench, EvalCrafter, and VBench-Long measure perceptual quality, prompt alignment, motion, and long-video consistency \citep{huang2024vbench,liu2024evalcrafter,huang2024vbenchpp}. StoryEval, StoryBench, and ViStoryBench add event completion, sequential tasks, or story-visualization criteria \citep{wang2025storyeval,bugliarello2023storybench,vistorybench2025}, while NarrLV and LoCoT2V-Bench explicitly evaluate narrative richness, event-level alignment, temporal consistency, and higher-level narrative realization \citep{feng2026narrlv,zheng2026locot2vbench}. MSVBench, EntityBench, and NarraStream-Bench specialize in multi-shot coherence, long-range entity consistency, or continuity across evolving prompts \citep{msvbench2026,he2026entitybench,liu2026narrastream}. MSAVBench and LongAV-Compass broaden evaluation to audio-visual generation and synchronization \citep{msavbench2026,liu2026longavcompass}; DirectorBench and FilmBench provide directorial, personalized, or professional cinematic diagnostics \citep{directorbench2026,wang2026filmbench}. SLVMEval instead meta-evaluates whether long-video evaluators detect controlled degradations, indicating that evaluator reliability must itself be tested \citep{matsuda2026slvmeval}.

These benchmarks are complementary rather than interchangeable. CineScope-Metric does not replace their specialized perceptual, audio-visual, entity-consistency, or cinematic taxonomies. It is distinguished by a compact final-output protocol centered on four episode-level Global dimensions and twenty human-aligned criteria. Production-trajectory localization belongs to the review module within CPPE and is evaluated separately from CineScope-Metric.

\subsection{Self-Improving and Agentic Video Generation}

Several recent systems explicitly frame video generation as agentic or self-improving. AesopAgent studies an agent-driven evolutionary system for story-to-video production \citep{wang2024aesopagent}. VideoAgent emphasizes self-improving video generation through iterative agent feedback \citep{soni2024videoagent}. VISTA performs test-time self-improvement by refining prompts with self-evaluation and multi-agent critique for a current generation request \citep{long2025vista}. A$^2$RD formulates long-video synthesis as a closed-loop agentic autoregressive diffusion process with consistency enforcement \citep{long2026a2rd}, while VideoWeaver evaluates and evolves reusable skills for agentic long-video generation \citep{wei2026videoweaver}. These works are closest in motivation and justify a precise boundary for our contribution. CPPE differs in the optimization object and deployment criterion: it does not primarily optimize a single test-time prompt, autoregressive diffusion trajectory, current story output, or reusable skill alone. Instead, it converts recurring failures across productions into typed persistent policy patches, admits them through fresh paired replay and risk accounting, and evaluates transfer on unseen stories under a frozen policy.

\subsection{Self-Refinement, Reflection, and Agent Memory}

Self-refinement and reflection methods let language agents critique their own outputs, store feedback, and condition future attempts on past experience. Reflexion, Self-Refine, Voyager-style skill libraries, and related memory-based agents show that textual feedback can improve agent behavior without gradient updates \citep{shinn2024reflexion,madaan2023selfrefine,wang2023voyager,zhao2024expel}. Our setting differs in two ways. First, we separate \emph{in-run repair} from \emph{across-run rule evolution}: a fixer may repair the current artifact, but only recurring failures that pass a recurrence test become candidates for persistent rule changes. Second, the write-back object is not an unconstrained memory entry; it is a typed patch over prompt rules, thresholds, genre profiles, or review policies, with risk levels and regression gates.

At the broader agent level, reusable skill, context, workflow, and harness optimization treats persistent external state as an editable object. SkillOpt evolves bounded skill documents, Agentic Context Engineering retains and revises structured context, and AgentX uses trace-derived semantic updates with replay \citep{yang2026skillopt,zhang2026ace,lao2026agentx}. ADAS and AFlow automate agent or workflow design \citep{hu2024adas,zhang2024aflow}, while TTHE, AutoHarness, DemoEvolve, RHO, AHE, NLAH, and HarnessFix further study persistent test-time or harness adaptation \citep{nie2026tthe,lou2026autoharness,che2026demoevolve,pan2026rho,lin2026ahe,pan2026nlah,chen2026harnessfix}. CPPE applies this specialization to long-horizon video production by tying bounded text-space policy patches to typed production traces, deterministic structural guards, and unseen-story transfer.

\subsection{Prompt, Program, and Text-Space Optimization}

Automatic Prompt Engineer, ProTeGi/APO, OPRO, TextGrad, DSPy, and related systems treat natural-language prompts or LM programs as optimizable objects \citep{zhou2023ape,pryzant2023automatic,yang2024opro,yuksekgonul2024textgrad,khattab2023dspy,opsahlong2024mipro,agrawal2026gepa,han2025mapgd}. We share their premise that text-space parameters can be improved without updating model weights. The difference is the deployment substrate and safety constraint. We optimize a multi-stage production policy whose failures are observed on explicit intermediate states, not a single prompt scored on a fixed training set. Candidate edits must be bounded, typed, auditable, and regression-tested because a prompt-rule change can affect story decomposition, shot planning, spatial consistency, safety rewriting, and prompt rendering across future productions.

\subsection{Safe Improvement and Regression-Tested Repair}

Our deployment rule is closer in spirit to safe policy improvement and regression-tested program repair than to unconstrained prompt search \citep{kakade2002approximately,pirotta2013safe,thomas2015high,laroche2019spibb,legoues2019automated}. A candidate rule patch is accepted only when replayed structural metrics do not regress on a held-out regression set; patches that can affect rendered semantics additionally require stochastic video-level comparison with a confidence margin. Recent analyses further caution that frequent harness updates do not by themselves establish downstream benefit and should be evaluated under fixed budgets and held-out tasks \citep{lin2026harnessbenefit,wang2026rethinkingharnesseval}. RHO \citep{pan2026rho} is a related but different line of agent optimization: it retrospectively optimizes general harness behavior from rollout self-preference rather than targeting long-horizon video generation. Our setting uses localized deterministic failures and paired counterfactual replay because video failures are delayed, multi-stage, and often partially deterministic before rendering. The patched object is not executable code and the environment is not a Markov decision process; it is a text-space policy controlling a compound generative workflow.

\subsection{Boundary of Novelty}

We do not claim a new video backbone, diffusion architecture, generic LLM optimizer, or open-ended self-modifying agent. The novelty lies in the combination required by long-horizon story-to-video production: explicit intermediate states that make structural failures measurable; cross-production case memory that abstracts local failures into recurring patterns; typed persistent patches over a multi-stage text-space policy; and fresh replay gates that turn adaptive self-evolution into a certifiable sequence of risk-adjusted structural improvements. Table~\ref{tab:rw-comparison} summarizes this boundary against planning agents, reflection systems, prompt optimizers, and within-request video refinement agents.

\begin{table}[t]
\centering
\small
\caption{Positioning relative to adjacent method families; ``persistent update'' denotes changing future workflow behavior rather than only repairing the current output.}
\label{tab:rw-comparison}
\resizebox{\linewidth}{!}{%
\begin{tabular}{lccccc}
\toprule
Method family & Long-horizon video & Cross-production update & Typed policy patch & Fresh replay gate & Transfer test \\
\midrule
Story/video planning agents & \checkmark & - & - & - & limited \\
Within-request video refinement agents & \checkmark & - & prompt/output & critique loop & current request \\
Prompt / LM-program optimizers & optional & \checkmark & prompt/program & validation & task split \\
Reflection / memory agents & optional & memory & free-form & - & task-dependent \\
\textbf{CineForge} & \checkmark & \checkmark & bounded policy patch & \checkmark & unseen stories \\
\bottomrule
\end{tabular}%
}
\end{table}

\section{Additional Experimental and Method Details}
\label{app:appendix}

\subsection{Complete Experimental Setup and Protocols}
\label{app:experimental-protocols}

\subsubsection{Data Construction and Split Allocation}
\label{app:cineverse}

We construct \textbf{CineScope-Data} to better match long-video production requirements and to provide evaluation-ready source material in which cross-scene dependencies, pacing variation, character development, and state continuity can be meaningfully tested. The suite contains 100 long-form benchmark scripts covering diverse genres, narrative organizations, character configurations, temporal spans, and scene-transition patterns. Each item contains a complete script or script-like narrative and a stable identifier from which task packages are derived. All 100 CineScope-Data scripts are used: 60 belong to the evolution-collection split, and the remaining 40 join 40 AnimeShooter and 40 ViStoryBench scripts to form a 120-script non-collection pool. Ten scripts from each source form the 30-script human-alignment split, 15 from each form the 45-script replay-admission split, and the remaining 15 from each form the disjoint 45-script system-test split.

\textbf{Corpus provenance and release.} CineScope-Data draws on domestic and international long-form novels, stories, and screenplays that are publicly accessible or used under explicit authorization. To avoid overstating provenance before the final corpus audit, individual platforms are not named here. Upon release, a dataset card and corpus manifest will disclose item-level provenance, language, license status, inclusion and exclusion decisions, deduplication and normalization procedures, and the identifiers assigned to every experimental split.

\textbf{Data sources and split roles.} We map CineScope-Data, AnimeShooter \citep{qiu2025animeshooter}, and ViStoryBench \citep{vistorybench2025} to a common task-package and trace schema. Table~\ref{tab:cinescope-data} records the source-level allocation; the released manifest records the exact story identifiers, the three disjoint 20-script round subsets that exhaust the CineScope-Data evolution-collection split, the 30 human-alignment identifiers and their development/held-out roles, the three disjoint 15-script replay-admission sets, and the fixed 45-script system-test identifiers. The evolution-collection, human-alignment, replay-admission, and system-test splits are mutually disjoint. Within each experimental track, all outputs sharing a \texttt{story\_id} remain within the same declared split. Shot- or scene-level evidence is aggregated within its parent story.

\begin{table}[htbp]
\centering
\scriptsize
\caption{Source allocation for the reported experiments. The 60-script evolution-collection split supports case formation and candidate compilation; the 30-script human-alignment split supports metric calibration and held-out alignment; the 45-script replay-admission split supports only fresh candidate admission; and the disjoint 45-script system-test split supplies system comparisons and every unseen measurement.}
\label{tab:cinescope-data}
\resizebox{\linewidth}{!}{%
\begin{tabular}{lccccc}
\toprule
Text source & Total scripts & Evolution collection & Human alignment & Replay admission & System test \\
\midrule
CineScope-Data & 100 & 60 & 10 & 15 & 15 \\
AnimeShooter \citep{qiu2025animeshooter} & 40 & - & 10 & 15 & 15 \\
ViStoryBench \citep{vistorybench2025} & 40 & - & 10 & 15 & 15 \\
\midrule
Total & 180 & 60 & 30 & 45 & 45 \\
\bottomrule
\end{tabular}%
}
\end{table}

\textbf{Experimental tracks.} CineScope-Metric alignment uses the dedicated 30-script human-alignment split (10 per source; five development and five held-out stories per source) and renders each script twice with CineForge-Produce, MovieAgent, and AniMaker, yielding 180 human-study videos; development ratings select the metric, and held-out stories evaluate the frozen configuration. Both alignment roles are system-unseen and disjoint from CineForge policy development, CPPE collection, replay admission, stopping, and all subsequent system comparisons. Their generation and evaluation are performed at read-only checkpoints, and no resulting artifact is written to the evolving system. The generator comparison evaluates CineForge-Produce, ViMax, AniMaker, and MovieAgent on the fixed 45-script system-test split. The controlled optimization comparison uses the same 60-script evolution-collection split for each optimization method and tests the resulting systems on that 45-script system-test split; CineForge-Evolve additionally uses the separate 45-script replay-admission split solely through its declared candidate gate. The round-wise self-evolution experiment uses three disjoint 20-script subsets that exhaust the CineScope-Data evolution-collection split; at each checkpoint, seen measurements use the corresponding 20-script round subset, whereas unseen measurements use the fixed 45-script system-test split. AnimeShooter and ViStoryBench do not enter collection or seen measurements, but 15 scripts from each contribute to every unseen measurement through the system-test split. All human-alignment and system-test identifiers and standardized evaluation outputs are excluded from collection, CaseMemory, patch compilation, replay admission, checkpoint selection, and stopping. Replay-admission identifiers and artifacts are excluded from collection, CaseMemory, pattern induction, patch compilation, final system testing, and stopping; they are consumed only by the predeclared candidate-admission gate.

\subsubsection{Unified Video-Production and Automatic-Evaluation Protocol}

\textbf{Shared video-production and video-evaluation configuration.} All compared production systems, including CineForge and all baseline or adapted variants, use the same text and video backbones: GLM~5.1 for text-side production calls and Seedance~2.0 for video generation. CineScope-Metric uses Qwen3.6-Plus for all visual-language evaluation calls (API model ID: \texttt{qwen3.6-plus-2026-04-02}); this VLM configuration is frozen across all system comparisons. The production systems receive the same source material, generation budget, output specification, and concatenation/post-processing policy, so each comparison isolates the production framework or optimization mechanism rather than the underlying backbones. The shared configuration uses temperature \(0.7\), top-\(p\) \(0.9\), an 8192-token maximum output length, at most three retries, and a concurrency cap of four; exposed API seeds are recorded. Video requests use a 16:9 aspect ratio, 1080p resolution, and 24\,fps. The run manifest records these settings and any interface-specific fields exposed by each provider.

\textbf{Repeated-generation and repeated-evaluation protocol.} For every story-system pair in the automatic-evaluation tracks, we generate two videos with independent generation randomness under identical model, budget, and input settings. Each generated video is evaluated twice independently by each automatic evaluator, producing $2\times2=4$ call-level scores per evaluator. For each evaluator separately, these four scores are averaged equally to obtain the unique story-system score used in every subsequent comparison and statistical analysis under that evaluator. Review LLM calls and Review issues from the globally frozen test reviewer are averaged over the two production runs. Replicate identifiers, exposed seeds, evaluator-call identifiers, invalid-run handling, and the four-to-one aggregation are retained in the run manifest; statistical analyses treat the story, not an individual video or evaluator call, as the independent unit.

\subsubsection{Human-Alignment Protocol}

\textbf{Human-study protocol.} Ten trained evaluators recruited from a university score complete videos using the same 20 definitions and 0-5 anchors as CineScope-Metric; each of the two generated videos per story-generator pair receives at least five independent ratings balanced by source, generator, and generation replicate. Human labels validate CineScope-Metric and are not collected for VBench-Long or ScriptAgent. Ratings are first aggregated per video and then across the two generation replicates, while the complete story remains the independent statistical unit.

\textbf{Human-evaluation configuration.} We recruited ten evaluators from a university and trained them on the narrative-video rubric and qualification examples before formal scoring. Before participation, evaluators received a description of the research purpose, tasks, expected time, compensation, and data use and provided informed consent; participation was voluntary and withdrawal was permitted at any time. Assignments were balanced and randomized, and system identities were hidden. Records use anonymous rater identifiers, omit public personally identifying information, and are stored under access control. Missing responses, failed attention checks, and qualification failures are handled by predeclared exclusion and replacement rules rather than post-hoc score inspection.

\textbf{Evaluator calibration.} Development-set human ratings are used to select temporal-scale weights, within-Global aggregation weights, confidence and abstention thresholds, and bounded revisions to the evaluator prompts. Selection maximizes development alignment subject to stable score validity and evidence coverage; equal weights and the previous prompt version are retained when a candidate does not improve the preregistered development objective. All weights, gates, prompts, retry behavior, and model versions are frozen before any held-out score is inspected, and held-out ratings never trigger further evaluator changes.

\textbf{Human-rater audit.} The released annotation audit records rater qualifications, the number of raters assigned to each video, repeated assignments, valid-rating coverage, inter-rater reliability, adjudication rate, invalid responses, and median annotation time per episode. Qualification requires experience with narrative video or script analysis, completion of the CineScope-Metric rubric tutorial, and a held-out calibration batch. CineScope-Metric annotations concern visible final-video evidence; production-stage cause labels are collected separately for CPPE process audits and do not change the human benchmark score.

\textbf{Alignment statistics.} Human consensus is averaged over valid raters per video and subdimension; Global scores are first aggregated within rater and then averaged, and the two generation replicates are averaged for each story-generator pair. For each analogous machine scope or subdimension, the two generation replicates and two evaluator calls per replicate produce four call-level scores, which are averaged equally to obtain one story-generator score before any downstream analysis. Spearman correlation is computed across story-generator units for each subdimension, Global aggregate, and optional overall aggregate. Bradley-Terry agreement uses declared tie handling on within-story comparisons among the three matched generators after replicate aggregation. Scope-level Spearman uses Fisher-\(z\) macro-aggregation and BT uses valid-pair weighting. For the held-out results in Table~\ref{tab:cinescope-validation}, source-stratified cluster bootstrap resamples the 15 held-out \texttt{story\_id}s with all generation replicates, machine calls, and raw ratings, refitting BT per bootstrap replicate. Dimension-wise tests use Benjamini-Hochberg FDR correction.

\subsubsection{External-Evaluator Protocol}

Because complete episodes may exceed the native input horizon of the companion evaluators, VBench-Long and ScriptAgent use frozen input-only long-video wrappers for video decoding, temporal partitioning, coverage tracking, and aggregation of supported evaluation units. These wrappers preserve each evaluator's original metric definitions, judging criteria, and score semantics; wrapper instructions only expose partitioned inputs and add no CineScope-Metric dimensions, production traces, or optimization feedback. VBench-Long otherwise retains its published implementation. ScriptAgent additionally uses a frozen source/script-and-video field adapter and is excluded from case generation, patch admission, checkpoint selection, and stopping. Every automatic evaluator is called twice independently on each of the two generated videos, and, for each evaluator separately, the resulting four scores are averaged equally to obtain the unique story-system result described above. The run manifest records exact model and prompt versions, input modalities, reference-audio policy, temporal sampling, evaluator-call identifiers, aggregation, retries, seeds when exposed, invalid-output handling, and cost. Each baseline receives only fields supported by its documented interface, and retained and omitted fields are recorded by the frozen adapter. The adapters add neither manually authored semantic annotations nor privileged traces; unavoidable interface mismatches are declared and analyzed separately.

\subsubsection{Supplementary Generator-Evaluation Results}

Table~\ref{tab:generator-comparison-companion} reports the VBench-Long and ScriptAgent panels corresponding to the same generator comparison and test set as Table~\ref{tab:generator-comparison}. VBench-Long measures complementary perceptual properties, whereas ScriptAgent provides an additional story-level critic view; neither evaluator is combined with CineScope-Metric or used as optimization feedback.

\begin{table}[htbp]
\centering
\scriptsize
\caption{Supplementary panels for the generator comparison in Table~\ref{tab:generator-comparison}: (b) VBench-Long and (c) ScriptAgent CriticAgent results on the same test set. Bold marks the best result in each column.}
\label{tab:generator-comparison-companion}
\textbf{(b) VBench-Long}\\[2pt]
\begin{tabular*}{\textwidth}{@{\extracolsep{\fill}}lcccccc}
\toprule
Generator & SC $\uparrow$ & BC $\uparrow$ & MS $\uparrow$ & DD $\uparrow$ & AQ $\uparrow$ & IQ $\uparrow$ \\
\midrule
ViMax \citep{huang2026vimax} & 0.9537 & \textbf{0.9638} & \textbf{0.9892} & 0.3914 & 0.6144 & 0.7574 \\
AniMaker \citep{shi2025animaker} & \textbf{0.9548} & 0.9576 & 0.9824 & \textbf{0.7054} & 0.6243 & \textbf{0.7746} \\
MovieAgent \citep{wu2025movieagent} & 0.9145 & 0.9519 & 0.9876 & 0.6368 & 0.5923 & 0.7057 \\
\midrule
CineForge-Produce & 0.9243 & 0.9531 & 0.9885 & 0.4185 & \textbf{0.6495} & 0.7466 \\
\bottomrule
\end{tabular*}
\medskip

\textbf{(c) ScriptAgent CriticAgent}\\[2pt]
\begin{tabular*}{\textwidth}{@{\extracolsep{\fill}}lcccccc}
\toprule
Generator & Critic avg. $\uparrow$ & CCA $\uparrow$ & KBL $\uparrow$ & VDF $\uparrow$ & EAM $\uparrow$ & NPT $\uparrow$ \\
\midrule
ViMax \citep{huang2026vimax} & 3.344 & 3.119 & 3.030 & 3.785 & 3.322 & 3.467 \\
AniMaker \citep{shi2025animaker} & 3.380 & 3.080 & 2.887 & 3.900 & 3.433 & 3.600 \\
MovieAgent \citep{wu2025movieagent} & 3.785 & 3.575 & \textbf{3.664} & 3.875 & \textbf{4.089} & 3.721 \\
\midrule
CineForge-Produce & \textbf{3.913} & \textbf{3.821} & 3.483 & \textbf{4.389} & 3.934 & \textbf{3.936} \\
\bottomrule
\end{tabular*}
\end{table}

\subsubsection{Self-Evolution Evaluation Protocol}
\label{app:eval}

Three consecutive evolution rounds update the policy from Round~0 through Round~3. For round $g$, the multi-story collection set $D_g$ and current memory $M_{g-1}$ induce recurring patterns $\mathcal B_g$ and compiled candidates $\mathcal C_g$ before fresh replay is revealed:
\[
\begin{aligned}
(\mathcal B_g,\mathcal C_g,M_g)
&=\operatorname{Compile}\!\left(H^{(g-1)};D_g,M_{g-1}\right),\\
\Delta_g^\star
&=\operatorname{Admit}\!\left(H^{(g-1)};\mathcal C_g,R_g^{\mathrm{det}},R_g^{\mathrm{vis}}\right),\\
H^{(g)}&=\begin{cases}H^{(g-1)}\oplus\Delta_g^\star,&\Delta_g^\star\ne\varnothing,\\H^{(g-1)},&\Delta_g^\star=\varnothing.\end{cases}
\end{aligned}
\]
For the reported sequence, $R_g$ is round $g$'s fresh 15-script replay-admission set, containing five scripts from each source; $R_1,R_2,R_3$ are mutually disjoint and are revealed only after $\mathcal C_g$ and its selection rule are frozen. Structural admission uses $R_g^{\mathrm{det}}=R_g^{\mathrm{diag}}\cup R_g$, where diagnostic failures are additional fixed checks and do not count toward the fresh replay sample. For a stochastic-impact candidate, the same story identifiers define $R_g^{\mathrm{vis}}=R_g$, evaluated through fresh independent paired generation and scoring replicates. Neither replay identifiers nor their artifacts can update $M_g$, induce patterns, compile candidates, enter final system testing, or influence stopping; they are consumed only by the candidate-admission gate.

Once \(H^{(g)}\) is immutable, seen and unseen measurements are generated in a stateless, write-blocked copy of that checkpoint. The seen measurement reuses only the corresponding collection story identifiers; its newly generated videos and all measurement artifacts are excluded from $M_g$ and later rounds. The unseen measurement uses the separate fixed 45-script system-test split \(U\). The three-round allocation and stopping rule are hash- and time-stamped before any checkpoint report is unsealed, and unseen reports are revealed only after Round~3 is frozen. No task, video, trajectory, review finding, validator record, CineScope-Metric score, ScriptAgent score, or standardized review-audit output from \(U\), and no post-round measurement artifact from the seen set, is available to any self-evolution component.

\textbf{Globally frozen review-workload audit.} Let \(M_{\mathrm{audit}}\) denote a reviewer snapshot fixed before Round~0 is evaluated. It includes the reviewer model and prompts, call-trigger thresholds, issue taxonomy and decision thresholds, routing, deduplication, retry policy, and maximum Review-Rewrite attempts. For every seen and unseen checkpoint test, the operational reviewer fields of \(H^{(g)}\) are temporarily overridden by \(M_{\mathrm{audit}}\) inside the write-blocked evaluation copy; the override, generated trajectory, findings, repair records, and scores are discarded after measurement. The next collection/evolution round resumes from the unchanged policy \(H^{(g)}\), whose production thresholds and review routes remain editable. Review calls count production-trajectory review and fixer invocations triggered under \(M_{\mathrm{audit}}\), and Review issues follow its fixed deduplication policy. Both exclude initial generation, post-hoc CineScope-Metric/ScriptAgent evaluation, pattern compilation, and replay. The manifest records the snapshot hash, verifies that it is identical across Round~0--Round~3, and logs that every CaseMemory, policy, threshold, prompt, and replay-pool write interface was disabled during evaluation.

\subsection{Detailed CineScope-Metric Protocol}
\label{app:cinescope-details}

\textbf{Benchmark item.} Let \(x\) be the source narrative, \(r\) the permitted task package, and \(V=\{V_1,\ldots,V_n,V^{\mathrm{epi}}\}\) the generated shots and the episode. A CineScope-Metric item is $q=(x,r,V)$. The metric receives the same permitted source materials and final rendered output for every generator; it does not receive the private production trajectory $Z$, CPPE review findings, evidence memory, candidate rules, or replay decisions. This input boundary makes CineScope-Metric independent of the optimization mechanism.

\textbf{Score and eligibility contract.} Human raters and the VLM use the same integer scale \(\{0,1,\ldots,5\}\) for every subdimension. Zero denotes complete failure or contradiction only when the subdimension is eligible at the current temporal scope; the absence of assessable evidence in an ineligible window produces exclusion or abstention, not a zero. Five denotes fully consistent support, with subdimension-specific anchors for levels 1-4. Before emitting a score for a 150-second or 30-second window, the evaluator first returns an eligibility decision based on (i) whether the source-text evidence aligns with that video window and (ii) whether the current subdimension can be judged from the available scope. Only eligible windows are scored. The eligibility contract, prompts, and thresholds are frozen before held-out evaluation and are independent of generator identity and the eventual score. For every scored subdimension and temporal scope, the evaluator returns a concise description, supporting or conflicting video timestamps, confidence, and abstention. An evaluation that yields a score without video-grounded evidence is invalid and is rerun under the frozen retry policy.

\textbf{Twenty-item rubric inventory.} Each Global dimension aggregates five directly scored subdimensions:
\begin{itemize}[leftmargin=*,itemsep=1pt,topsep=2pt]
    \item \textbf{Causal State (CS):} key-event visibility; temporal-order correctness; causal-chain readability; state-change continuity; consequence follow-through and closure.
    \item \textbf{Directorial Orchestration (DO):} shot-function match; visual-focus clarity; plot-serving shot scale and camera movement; spatial-blocking continuity; visual emphasis and beat hierarchy.
    \item \textbf{Pacing and Resource Allocation (PR):} importance-aware duration allocation; key-beat elaboration; information-density comfort; tension curve and editing rhythm; closure and breathing room.
    \item \textbf{Character Arc (CA):} character-identity recognizability; goal and motivation traceability; behavioral and emotional progression; relationship-dynamics progression; arc resolution and individuation.
\end{itemize}

\textbf{Multiscale aggregation.} CineScope-Metric scores the complete episode and applies the eligibility gate above to aligned, non-overlapping 150-second windows \(\mathcal{W}_{150}\) and 30-second windows \(\mathcal{W}_{30}\); a final shorter window is retained and duration-weighted when eligible. For criterion $k$, window $w$, and scale $t\in\{150,30\}$, let $a_{k,w}\in\{0,1\}$ be the frozen eligibility decision and $d_w$ the window duration. Define
\[
\bar{s}_{k,t}=\frac{\sum_{w\in\mathcal W_t}a_{k,w}d_ws_{k,w}}
{\sum_{w\in\mathcal W_t}a_{k,w}d_w},\qquad
I_{k,t}=\mathbf 1\!\left\{\sum_{w\in\mathcal W_t}a_{k,w}d_w>0\right\},
\]
with $\bar{s}_{k,t}$ omitted when $I_{k,t}=0$. The final score renormalizes the frozen temporal weights over available evidence:
\[
s_k=\frac{\alpha_{\mathrm{epi}}s_{k,\mathrm{epi}}+
\sum_{t\in\{150,30\}}\alpha_t I_{k,t}\bar{s}_{k,t}}
{\alpha_{\mathrm{epi}}+\sum_{t\in\{150,30\}}\alpha_t I_{k,t}},
\qquad \sum_t\alpha_t=1,
\]
\[
s_g=\sum_{k\in K_g}\beta_{gk}s_k,
\qquad \sum_{k\in K_g}\beta_{gk}=1,
\]
where \(K_g\) contains the five subdimensions of Global dimension \(g\). Long-range criteria such as arc resolution and causal closure therefore rely primarily on complete-episode evidence unless an aligned window contains a criterion-relevant milestone; a locally absent ending is never treated as a global failure. Temporal weights \(\alpha\), within-Global weights \(\beta\), eligibility rules, prompts, abstention behavior, and retries are calibrated only on evaluator-development stories and then frozen. Equal weights are the preregistered fallback when calibration does not improve development alignment. All four Global scores and all 20 subdimension scores, together with per-system criterion-by-scale inclusion, exclusion, and abstention rates, are released; their overall average is used only as a compact summary.

\textbf{Metric ownership.} VBench-Long independently reports subject consistency, background consistency, motion smoothness, dynamic degree, aesthetic quality, and imaging quality \citep{huang2024vbenchpp}. These remain attributed to VBench-Long, are not passed through the CineScope-Metric judge, and are never averaged with CineScope-Metric or ScriptAgent. Unsupported dimensions are reported as N/A rather than zero.

\subsection{Detailed Production-Policy Components and Risk Levels}
\label{app:policy-components}

This subsection specifies which production-policy components CineForge-Evolve may update and the evidence required for deployment. Evolution is restricted to bounded, stage-local targets; higher-risk or system-level changes require stricter replay, human approval, or remain outside the automatic update boundary.

\begin{table}[h]
\centering
\scriptsize
\caption{Expanded production-policy components and update permissions.}
\begin{tabularx}{\linewidth}{>{\raggedright\arraybackslash}p{0.18\linewidth}X>{\raggedright\arraybackslash}p{0.20\linewidth}X}
\toprule
Component & Example & Evolution & Replay evidence \\
\midrule
Typed state schema & atoms, shot skeletons, scene-state ledgers, prompt records & no & schema validity \\
Stage prompt/contract & skeleton planner, narrator, renderer templates & yes & structural validators on replayed states \\
Production threshold/config & duplicate-dialogue threshold, retry limit & yes & paired metric difference and config diff \\
CPPE controls & recurrence $k_g$, opportunity/replay definitions, margins, confidence budgets & no within a sequence; human-approved recalibration starts a new one & calibration/approval audit and new sequence ID \\
Review rubric & CP issue definitions and repair routing & bounded & issue localization and old/new replay \\
Genre/style profile & style anchor, camera profile, lighting profile & bounded & held-out transfer and side effects \\
Deterministic validators & atom coverage, speaker-in-shot, scene-reference validity & manual only & validator regression tests \\
Global stage graph & stage order and mandatory checkpoints & manual only & integration tests and human approval \\
Model/API backend & LLM, image model, video generation service & no & environment report \\
\bottomrule
\end{tabularx}
\end{table}

\subsection{CPPE Structural Validators and Synthetic Replay Support}
\label{app:cppe-structural-validators}

\textbf{Deterministic structural metrics.} CPPE validators operate separately from rendered-outcome evaluators. For atom set \(\mathcal{A}\), coverage map \(\phi\), and scene blueprint \(\mathcal{B}_s\), the fixed replay audit includes
\[
\text{Cov}=\tfrac{|\{a:\exists s,\phi(a)=s\}|}{|\mathcal{A}|},\quad
\text{Dup}=\tfrac{\sum_a\max(0,|\phi^{-1}(a)|-1)}{|\mathcal{A}|},\quad
\text{JSONValid}=\tfrac{\#\text{schema-valid units}}{\#\text{units}},
\]
\[
\begin{aligned}
\text{HardPass}&=\tfrac{\#\text{units passing all hard rules}}{\#\text{units}},&
\text{Spatial}/\text{Facing}&=\tfrac{\#\text{shots consistent with }\mathcal{B}_s}{\#\text{shots}},\\
\text{BadCase}&=\tfrac{\#\text{units with a critical/major issue}}{\#\text{units}}.&&
\end{aligned}
\]
These checker outputs are exactly reproducible only when the affected path after the frozen cache boundary is deterministic, as specified in Section~\ref{sec:formulation}; otherwise the patch is routed to stochastic admission. MovieNet \citep{huang2020movienet}, MovieBench \citep{wu2025moviebench}, Book2Movie \citep{tapaswi2015book2movie}, and a Chinese public-domain set may support supplementary structural transfer analysis but are not members of the paired CineScope-Metric human subset.

\textbf{Synthetic replay support.} A synthetic CPPE review/replay suite provides generated stories, known atom sets, injected omissions, duplications, schema errors, spatial contradictions, and ground-truth structural labels. It tests trajectory localization and recurrence behavior rather than CineScope-Metric human alignment.

\subsection{Replay Certificate and Audit Conditions}
\label{app:replay-conditions}

The sequential certificate applies only when the complete candidate class and its selection rule are fixed before replay-admission stories are revealed. Replay \texttt{story\_id}s cannot contribute to case memory, pattern induction, or patch compilation in the same round. The candidate-batch hash, class size, compile time, and replay-reveal time are stored in the audit, and at most one patch is selected per round; all remaining candidates are invalidated for that round. An exact comparison is permitted only after a declared cache boundary whose affected downstream transformations are deterministic. The cache contains frozen upstream artifacts, task inputs, and validator versions, but never hides an editable prompt, template, or rule inside the fixed context. Patches that require fresh LLM or video calls use stochastic admission even when provider seeds are available. Validator definitions, recurrence thresholds and opportunity models, admission margins and confidence budgets, and the weights in structural welfare, guard debt, and dependency-confinement change are frozen before the certified sequence begins. Any change to these CPPE controls requires independent human approval and held-out recalibration, starts a new certified sequence with a new identifier and confidence allocation, and cannot inherit the prior certificate.

For low-risk structural deployment, every protected deterministic metric uses zero tolerance: a candidate with any negative protected delta is rejected. A nonzero tolerance for a higher-risk target must be predeclared, reported per metric, and approved under the corresponding risk gate. Every \texttt{stochastic-impact} candidate additionally uses repeated paired generation or rating under the frozen CPPE admission configuration and the preregistered confidence margin in Proposition~\ref{prop:stochastic-gate}. CineScope-Metric, VBench-Long, ScriptAgent, human preference, and final held-out video outcomes remain independent empirical reports rather than admission signals.

The audit for each accepted and rejected patch reports target improvement, every guard delta, dynamic-slice support, weighted off-slice change, replay sample size, candidate-class size, and acceptance margin. The certificate is tied to the frozen target distribution and replay construction; a material domain shift requires a new replay pool and does not inherit the earlier guarantee.

\begin{table}[h]
\centering
\small
\caption{Patch risk levels.}
\begin{tabularx}{\linewidth}{lXXX}
\toprule
Risk & Editable target & Deployment gate & Example \\
\midrule
L0 & soft prompt phrasing & structural replay; visual gate if flagged & adjust narration wording \\
L1 & production threshold/config & replay, config diff; visual gate if flagged & tighten duplicate threshold \\
L2 & stage contract, genre profile, review rubric & replay, visual gate, impact review & add spatial-state constraint \\
L3 & safety rule or validator semantics & human approval required & change policy-violation rule \\
\bottomrule
\end{tabularx}
\end{table}

\subsection{Stochastic, Recurrence, and Transfer Guarantees}
\label{app:aux-theory}

The fixed-item structural gate in Section~\ref{sec:method} is exact only on deterministic replay paths; Theorem~\ref{thm:fresh-replay} separately provides a high-probability population statement. The following results cover bounded stochastic admission, recurrence, and held-out transfer.

\begin{proposition}[Sequential high-confidence gate for stochastic admission]
\label{prop:stochastic-gate}
At round $t$, assume each CPPE stochastic-admission metric is bounded in \([0,1]\), the sum of nonnegative weights is \(W_{\mathcal A}\), and every incumbent/candidate score is estimated from $n_{V,t}$ independent story-level paired evaluations. Within each story and policy, the two generation replicates and two evaluation calls per replicate are averaged equally before the candidate-incumbent difference is formed. If $K_t$ candidates are compared with one incumbent and $\delta_t^V>0$, define
\[
s_t=W_{\mathcal A}\sqrt{\frac{\log(2(K_t+1)/\delta_t^V)}{2n_{V,t}}}.
\]
If the candidate satisfies \(\widehat{\Delta A}_t\ge m_{V,t}+2s_t\), then with probability at least $1-\delta_t^V$ its expected admission-score improvement is at least $m_{V,t}$. Across adaptive rounds, if $\sum_t\delta_t^V\le\delta_V$, the statement holds simultaneously for every accepted stochastic-impact patch with probability at least $1-\delta_V$. This proposition concerns CPPE admission only and makes no CineScope-Metric claim.
\end{proposition}

Lemma~\ref{thm:recurrence} supplies the recurrence result under conditionally independent case opportunities; a union bound controls family-wise false triggers across monitored pattern families. Correlated operational streams may be calibrated empirically on a held-out no-update stream, but that procedure is not covered by the i.i.d. bound.

\begin{proposition}[Complete operational cause-level transfer decomposition]
\label{prop:cause-transfer}
Let \(\bar{\mathcal C}=\mathcal C\cup\{\texttt{unknown}\}\) be the frozen operational cause set. Under matched pre/post evaluation, let $r_c$ be the pre-update error mass of cause $c$ removed after the update and $a_c$ the post-update error mass of cause $c$ that was absent before the update. Then the exact held-out transfer gain is
\[
\Delta_{\mathrm{transfer}}
=\sum_{c\in\bar{\mathcal C}}(r_c-a_c)
=(r_{c^*}-a_{c^*})+\sum_{c\ne c^*}(r_c-a_c).
\]
Consequently, transfer is positive if and only if total removed error mass exceeds total newly introduced error mass. The earlier expression $\rho_{c^*}\eta_{c^*}-\sum_{c\ne c^*}\gamma_{c\leftarrow c^*}$ is a special case requiring no new target errors, no removed non-target errors, and unchanged unknown mass. An empirical claim additionally requires the lower endpoint of the story-bootstrap 95\% confidence interval for \(\Delta_{\mathrm{transfer}}\) to exceed zero.
\end{proposition}

\begin{table}[htbp]
\centering
\small
\caption{Theory-to-evidence map. Structural, stochastic, recurrence, and transfer claims require evidence collected under their separately stated conditions.}
\label{tab:theory-diagnostics-main}
\begin{tabularx}{\linewidth}{lXX}
\toprule
Theoretical object & What the theorem establishes & Evidence required for an empirical claim \\
\midrule
Dependency slice & changed deterministic fields are confined to the executed dependency closure & slice support, off-slice side-effect mass \\
Replay certificate & accepted patches improve future structural welfare after risk charges & replay margin, candidate class size, guard deltas \\
Sequential evolution & adaptive multi-round updates accumulate certified margins & Round~0--3 policy trajectories, accepted/rejected patch audit \\
Stochastic gate & round-allocated confidence margin controls bounded CPPE admission noise & repeated paired scores, candidate count, confidence radius \\
Recurrence control & recurrence threshold bounds false and missed triggers & no-update stream, opportunities, trigger counts \\
Cause-level transfer & total removed mass must exceed total newly added mass & per-cause removed/added errors, including unknown \\
Video utility & no structural implication; quality is an empirical outcome claim & CineScope-Metric-human alignment, CineScope-Metric/ScriptAgent seen-unseen scores \\
\bottomrule
\end{tabularx}
\end{table}

\subsection{Detailed Method Notes}
\label{app:method}

\textbf{CineForge-Produce: information-conserving decomposition.} A long story is decomposed into narrative atoms and downstream coverage maps. The structural check is completeness and non-duplication: every checkable atom should be assigned to at least one shot and should not be redundantly covered across incompatible units. This makes narrative preservation observable before video rendering.

\textbf{CineForge-Produce: deterministic scaffold and creative infill.} The production path fixes identifiers, schemas, and structural normalizers. LLMs fill bounded semantic fields such as shot intent, visual focus, and narration, while editable prompt-rendering templates remain part of $H$. Exact replay begins only after an explicit cache boundary and covers downstream deterministic transformations; any patch that requires creative infill or regeneration is handled by stochastic admission.

\textbf{CineForge-Produce: single-source-of-truth continuity.} Scene blueprints, state ledgers, force-vector-to-facing rules, and first-frame chains carry long-range continuity. When continuity failures recur, CineForge-Evolve targets the stage rule that derives the relevant state rather than patching a single prompt.

\textbf{CineForge-Produce: concrete production graph.} Table~\ref{tab:cineforge-produce-graph} summarizes the implemented story-to-video path. Each stage consumes versioned typed records, emits auditable artifacts, and runs a stage-appropriate validator before its outputs become eligible for downstream use.

\begin{table}[htbp]
\centering
\scriptsize
\setlength{\tabcolsep}{2.5pt}
\caption{Concrete CineForge-Produce graph. LLM and generative calls fill bounded creative fields; deterministic transforms preserve identifiers, dependencies, and validation records.}
\label{tab:cineforge-produce-graph}
\begin{tabularx}{\linewidth}{>{\raggedright\arraybackslash}p{0.16\linewidth}>{\raggedright\arraybackslash}p{0.18\linewidth}>{\raggedright\arraybackslash}p{0.20\linewidth}>{\raggedright\arraybackslash}X}
\toprule
Stage / owner & Typed input & Typed output & Operation and validation \\
\midrule
Narrative and episode planning / planner
& Task package: \texttt{story\_id}, source text, target budget, policy version
& Narrative atoms: \texttt{atom\_id}, event and state delta, episode assignment; episode plan
& An LLM fills bounded semantic fields; schema, identifier, completeness, and non-duplication checks validate the plan. \\
Scene and state planning / planner
& Atoms, episode plan, character and spatial state
& Scene blueprints, atom-coverage map, versioned character/spatial state ledgers
& LLM planning is followed by deterministic identifier and state propagation; causal, spatial, and coverage validators run at the boundary. \\
Shot design / director
& Scene blueprints, state ledgers, coverage map
& Shot contracts and storyboard records: \texttt{shot\_id}, intent, camera, timing, and dependencies
& The director call proposes shot-level creative fields; deterministic normalizers enforce required fields, timing, coverage, and facing constraints. \\
Character and scene assets / asset manager
& Shot contracts and versioned character/scene specifications
& Asset records, reference identifiers, and provenance links
& Asset calls instantiate reusable visual references; identity, specification, availability, and version checks gate their use. \\
Prompt rendering / renderer
& Shot contracts, asset identifiers, and templates in \(H\)
& Rendered prompt records and backend requests with reference and budget fields
& A deterministic template transform renders each request; contract, required-field, reference, and budget validators run before submission. \\
Image/reference generation / reference builder
& Rendered requests and reusable asset records
& Shot-specific first frames or reference images and a propagation manifest
& The image/reference backend produces visual anchors; accepted references propagate along the declared shot chain and are checked for identity and availability. \\
Video generation and local Review-Rewrite / generator and reviewer
& Rendered prompts, first-frame/reference chain, and generation settings
& Clip artifacts, local findings, repair attempts, and outcome lineage
& The video backend renders clips; bounded review and rewrite may revise the current artifact, while validity, event-completion, prompt-fidelity, and continuity checks are logged. \\
Composition and global review / composer and reviewer
& Accepted clips, timing plan, transitions, and episode order
& Composed episode, composition manifest, and global-review findings
& Deterministic concatenation and post-processing preserve order and timing; coverage, duration, causal, pacing, and long-range continuity checks audit the final episode. \\
\bottomrule
\end{tabularx}
\end{table}

All versioned plans, ledgers, assets, prompts, model requests and outputs, validator/reviewer findings, repair attempts, and composition records are persisted in the canonical trajectory \(Z\). A first frame or reference selected for a shot is recorded with its provenance and passed through the explicit shot dependency chain. For a proposed policy patch, the dynamic slice begins at the changed field and expands only through declared downstream dependencies. Paired replay fixes the task input, unaffected upstream artifacts, validator versions, and recorded exogenous conditions at the cache boundary, then recomputes deterministic descendants. Any slice that requires fresh LLM, image, or video generation is excluded from exact replay and evaluated through stochastic admission instead.

\textbf{CineForge-Evolve: production trajectory to evidence-grounded case.} CPPE inspects both the intermediate artifacts in the complete story-to-video trajectory and the rendered video evidence. Its multilevel discovery pass organizes checks into three evidence streams. The pipeline stream applies deterministic hard rules to schemas, identifiers, required fields, durations, and prompt contracts; checks local semantic propagation along beat-script-storyboard-prompt chains; inspects reusable character and scene assets; and audits cross-segment state, causal, transition, and duplication consistency. The local-video stream jointly reads the prompts and generated clips within a segment to detect prompt-fidelity, visual, motion, temporal, event-completion, local-continuity, and transition failures. The global-video stream compares the adapted episode script with the composed episode to inspect narrative alignment, completeness, coherence, progression and pacing, shot and cinematic organization, and long-range character and scene continuity.

Every detector emits a LocatedIssue $\ell$ rather than an immediate policy edit. It records what was observed, where it was observed, the supporting intermediate-artifact or video evidence, and a set of possible upstream causes. For video-side symptoms, the review module maps the observation to its clip and segment and traverses the production chain backward through prompt, storyboard/shot design, script, character/scene assets, and beat/episode planning. The assigned root-cause stage is the earliest stage at which the error is supported by the available evidence or at which a required constraint was first absent; a failure supported only by the rendered output remains attributed to the backend/provider, while insufficient or conflicting evidence yields \texttt{unknown}. This separation prevents the location where a defect becomes visible from being conflated with the stage that introduced it.

CPPE next matches duplicate LocatedIssues produced at different granularities, merges their pipeline, prompt/script, storyboard, local-video, and global-video evidence, and preserves the contributing records as lineage. The merged finding is routed by root-cause stage to a stage-specific critic, which normalizes the issue type and severity, confirms or abstains on the root cause, proposes the fix type and fix detail, and assigns a cluster key. The resulting standard issue $e$ therefore links an observed failure to evidence, an operational root cause, and a bounded repair proposal. Deterministic violations remain distinguished from reviewer hypotheses; neither type can authorize persistent deployment without the downstream replay gates.

\textbf{CineForge-Produce/Evolve interface.} A standard issue may trigger bounded Review-Rewrite for the current production and is always logged with its repair outcome. Fixed one-off findings remain auditable but do not support persistent evolution. Only unresolved or \texttt{stuck}, sufficiently confident, editable issues that recur across conditionally independent stories or predeclared opportunities enter CineForge-Evolve's case-to-pattern-to-policy path. The root-stage label is an auditable diagnostic hypothesis, whereas replayed target and guard changes provide the behavioral evidence for deployment. CineScope-Metric and the other outcome benchmarks are absent from discovery, diagnosis, repair, case accumulation, pattern induction, patch compilation, and admission.

\subsection{Algorithmic Details}
\label{app:algorithms}

\begin{algorithm}[htbp]
\caption{CineForge-Evolve (Case-to-Pattern-to-Policy Evolution). PatternInduction produces recurring-pattern set $\mathcal B_r$, CandidateCompilation produces candidate-patch set $\mathcal C_r$, the candidate set is frozen before fresh replay, and at most one accepted patch $\Delta_r^\star$ is deployed.}
\label{alg:cppe-evolve}
\small
\begin{algorithmic}[1]
\Require policy $H_r$, story collection $\mathcal D_r$, memory $\mathcal M_r$, replay pools
\Ensure updated policy $H_{r+1}$ and memory $\mathcal M_{r+1}$
\State $\mathcal M_{r+1} \gets \mathcal M_r$
\For{each story $x \in \mathcal D_r$}
    \State $(V,Z) \gets \Call{ProduceVideo}{H_r,x}$
    \State $\mathcal L \gets \Call{MultiLevelTrajectoryReview}{Z,V}$
    \State $\mathcal L \gets \Call{TraceRootCauses}{\mathcal L,Z,V}$
    \State $\mathcal L \gets \Call{DeduplicateAndMergeEvidence}{\mathcal L}$
    \For{each located issue $\ell \in \mathcal L$}
        \State $e \gets \Call{StageSpecificDiagnose}{\ell,Z,V}$
        \State $\mathcal M_{r+1} \gets \Call{StoreCase}{\mathcal M_{r+1},e,\mathit{repair\_outcome}}$
    \EndFor
\EndFor
\State $\mathcal B_r \gets \Call{PatternInduction}{\mathcal M_{r+1}}$
\State $\mathcal C_r \gets \Call{CandidateCompilation}{H_r,\mathcal B_r}$
\State $\Call{Freeze}{\mathcal C_r,\mathit{selection\_rule},\mathit{tie\_break}}$; $K_r \gets |\mathcal C_r|$
\State $\mathcal R_r \gets \Call{DrawFreshReplayAfterFreeze}{}$
\State $\mathcal R_{\mathrm{diag}} \gets \Call{TriggeringAndNearMissReplay}{\mathcal M_{r+1}}$
\State $\mathcal R_{\mathrm{det}} \gets \mathcal R_{\mathrm{diag}} \cup \mathcal R_r$
    \State $\mathcal R_{\mathrm{vis}} \gets \mathcal R_r$ \Comment{fresh paired replicates if stochastic-impact}
\State $\mathcal A_r \gets \varnothing$
\For{each $\Delta \in \mathcal C_r$}
    \State $H_{\Delta} \gets \Call{ApplyUpdate}{H_r,\Delta}$
    \State $g_{\mathrm{str}} \gets \Call{StructuralReplayGate}{H_r,H_{\Delta},\mathcal R_{\mathrm{det}}}$
    \State $g_{\mathrm{cert}} \gets \Call{CertifiedReplayMargin}{H_r,H_{\Delta},\mathcal R_r,K_r}$
    \State $g_{\mathrm{stoch}} \gets \neg\Call{StochasticImpact}{\Delta}$
    \If{\Call{StochasticImpact}{$\Delta$}}
        \State $g_{\mathrm{stoch}} \gets \Call{CPPEStochasticGate}{H_r,H_{\Delta},\mathcal R_{\mathrm{vis}},K_r,\delta_r^V}$
    \EndIf
    \State $\Call{RecordGateResultInAudit}{\Delta,g_{\mathrm{str}},g_{\mathrm{cert}},g_{\mathrm{stoch}}}$
    \If{$g_{\mathrm{str}} \land g_{\mathrm{cert}} \land g_{\mathrm{stoch}}$}
        \State $\mathcal A_r \gets \mathcal A_r \cup \{\Delta\}$
    \EndIf
\EndFor
\State $\Delta_r^\star \gets \Call{SelectAtMostOne}{\mathcal A_r,\mathit{selection\_rule},\mathit{tie\_break}}$
\If{$\Delta_r^\star$ exists}
    \State $H_{r+1} \gets \Call{ApplyUpdate}{H_r,\Delta_r^\star}$
\Else
    \State $H_{r+1} \gets H_r$
\EndIf
\State \Return $H_{r+1},\mathcal M_{r+1}$
\end{algorithmic}
\end{algorithm}

\begin{algorithm}[htbp]
\caption{CineForge-Produce (Invariant-Guided Generation). The procedure records coverage, scene state, prompts, and visual-anchor fields in the canonical trajectory $Z$.}
\label{alg:invariant-generation}
\small
\begin{algorithmic}[1]
\Require story $x$, production policy $H$, previous visual anchors $\mathcal M_v$
\Ensure episode $V$, typed trace $Z$, and coverage map $\phi$
\State $\mathcal A \gets \Call{ExtractNarrativeAtoms}{x}$
\State $\mathcal P \gets \Call{BuildEpisodeSceneShotPlan}{x,H}$
\State $\mathcal S \gets \Call{DeterministicScaffold}{\mathcal P,\mathcal A}$
\State $\phi \gets \Call{BuildCoverageMap}{\mathcal A,\mathcal S}$
\If{\Call{CoverageViolation}{$\phi$}}
    \State $\mathcal S \gets \Call{RepairCoverageDeterministically}{\mathcal S,\mathcal A,\phi}$
\EndIf
\For{each scene $s \in \mathcal S$}
    \State $B_s \gets \Call{BuildSceneBlueprint}{s,H}$
    \For{each shot $u$ in scene $s$}
        \State $\mathit{state}_u \gets \Call{DeriveSpatialState}{B_s,u}$
        \State $\mathit{text}_u \gets \Call{CreativeInfill}{u,\mathit{state}_u,H}$
        \State $\mathit{prompt}_u \gets \Call{RenderPrompt}{\mathit{text}_u,\mathit{state}_u,H}$
        \State $\mathit{clip}_u \gets \Call{GenerateVideo}{\mathit{prompt}_u,\mathcal M_v}$
        \State $\mathcal M_v \gets \Call{UpdateVisualAnchor}{\mathit{clip}_u}$
        \State $\Call{Add}{Z,u,\mathit{state}_u,\mathit{prompt}_u,\phi}$
    \EndFor
\EndFor
\State \Return $\Call{MergeClips}{\{\mathit{clip}_u\}},Z,\phi$
\end{algorithmic}
\end{algorithm}

\begin{algorithm}[htbp]
\caption{CineForge-Evolve Replay-Gated Policy Versioning. Deterministic hard guards apply to every patch, while stochastic-impact patches additionally undergo repeated paired admission.}
\label{alg:replay-gated-versioning}
\small
\begin{algorithmic}[1]
\Require policy $H$, patch $\Delta$, replay sets $\mathcal R_{\mathrm{det}}$ and $\mathcal R_{\mathrm{vis}}$, $K$, $\rho$, $\epsilon$, $m_V$, $\delta_t^V$
\Ensure policy $H_{\mathrm{new}}$ and decision $d$
\State $H_{\mathrm{cand}} \gets \Call{ApplyUpdate}{H,\Delta}$
\If{$\rho=L3 \land \neg\Call{HumanApproval}{\Delta}$}
    \State \Return $H,\textsc{RejectedByHumanGate}$
\EndIf
\State $\mathit{old} \gets \Call{RegressionTest}{H,\mathcal R_{\mathrm{det}}}$
\State $\mathit{new} \gets \Call{RegressionTest}{H_{\mathrm{cand}},\mathcal R_{\mathrm{det}}}$
\If{$\mathit{new}$ violates any hard constraint}
    \State $\Call{RecordRejectionInAudit}{\Delta,\textsc{HardConstraintFailure}}$
    \State \Return $H,\textsc{RejectedByHardRule}$
\EndIf
\If{any protected $\delta< -\epsilon[\rho]$}
    \State $\Call{RecordRejectionInAudit}{\Delta,\textsc{RegressionDrop}}$
    \State \Return $H,\textsc{RejectedByRegression}$
\EndIf
\If{\Call{OffSliceChange}{$H,H_{\mathrm{cand}},\mathcal R_{\mathrm{det}}$} exceeds $\mathit{gate}[\rho]$}
    \State $\Call{RecordRejectionInAudit}{\Delta,\textsc{UnattributedSideEffect}}$
    \State \Return $H,\textsc{RejectedBySideEffectGate}$
\EndIf
\If{\Call{StochasticImpact}{$\Delta$}}
    \State $\mathit{paired} \gets \Call{RepeatedCPPEAdmission}{H,H_{\mathrm{cand}},\mathcal R_{\mathrm{vis}}}$
    \If{$\mathit{paired}.\delta < m_V+2\Call{ConfidenceRadius}{K,\delta_t^V}$}
        \State $\Call{RecordRejectionInAudit}{\Delta,\textsc{StochasticGateFailure}}$
        \State \Return $H,\textsc{RejectedByStochasticGate}$
    \EndIf
\EndIf
\State $\Call{RecordAcceptanceInAudit}{\Delta,\mathit{old},\mathit{new}}$
\State \Return $H_{\mathrm{cand}},\textsc{Deployed}$
\end{algorithmic}
\end{algorithm}

\FloatBarrier
\subsection{Full Proofs}
\label{app:proofs}

\paragraph{Definitions and assumptions.}
For a story $x$ and declared cache boundary $b$, $c_b(x)$ contains fixed upstream artifacts, task inputs, and validator versions. Editable rules and templates remain in $H$. When all affected downstream transformations are deterministic, $T_H^d(x;c_b)$ denotes their trace. If a patch requires a stochastic model call, it is excluded from exact replay and handled by Proposition~\ref{prop:stochastic-gate}. The normalized structural welfare $Q(T)\in[0,1]$, validator set, guard weights $\lambda_j$, and trace-field weights $w_z$ are frozen before the certified sequence. With $q_j(T)\in[0,1]$ and $\Lambda=\sum_{j\in\mathcal G_\rho}\lambda_j$, define normalized guard debt
\[
D_{H,\Delta}(x)=
\begin{cases}
\Lambda^{-1}\sum_{j\in\mathcal G_\rho}\lambda_j
[q_j(T_H^d)-q_j(T_{H\oplus\Delta}^d)]_+, & \Lambda>0,\\
0, & \Lambda=0.
\end{cases}
\]
Thus $D_{H,\Delta}\in[0,1]$. The dependency-confinement term $U_{H,\Delta}\in[0,1]$ is defined below. For fixed \(\beta,\gamma\ge0\),
\[
Y_{H,\Delta}(x)=Q(T_{H\oplus\Delta}^d(x;c_b))-Q(T_H^d(x;c_b))
-\beta D_{H,\Delta}(x)-\gamma U_{H,\Delta}(x).
\]
Because the three terms are normalized, an explicit uniform bound is
\[
C=1+\beta+\gamma.
\]
If the deployment distribution changes materially, the replay pool and certificate must be rebuilt; Theorem~\ref{thm:fresh-replay} is stated for fixed \(\mathcal D\).

\paragraph{Cumulative structural-welfare form.}
Define
\[
\Delta Q_{0:K}:=
\mathbb E_{x\sim\mathcal D}Q(T_{H_K}^{d}(x;c_b))-
\mathbb E_{x\sim\mathcal D}Q(T_{H_0}^{d}(x;c_b)).
\]
On the simultaneous event in Theorem~\ref{thm:fresh-replay}, the accepted rounds \(\mathcal A\) satisfy
\begin{align*}
\Delta Q_{0:K}&\ge \sum_{r\in\mathcal A}m_r
+\beta\sum_{r\in\mathcal A}\mathbb E_{x\sim\mathcal D}D_{H_{r-1},\Delta_r}(x)
+\gamma\sum_{r\in\mathcal A}\mathbb E_{x\sim\mathcal D}U_{H_{r-1},\Delta_r}(x).
\end{align*}

\paragraph{Least fixed-point dependency closure.}
For each replay item \(x\), let \(V_x\) be the finite set of typed trace fields. Each deterministic transformation \(f_i\) logs input fields \(I_i(x)\subseteq V_x\) and output fields \(O_i(x)\subseteq V_x\). Let \(S_0(\Delta,x)\) be the fields directly written by the audited support of \(\Delta\). Define the monotone map
\[
\Phi_{x,\Delta}(S)=S\cup\{z\in V_x:\exists i,\exists y\in S\text{ such that }y\in I_i(x),\ z\in O_i(x)\}.
\]
Starting from \(S^{(0)}=S_0(\Delta,x)\), set \(S^{(r+1)}=\Phi_{x,\Delta}(S^{(r)})\). Since \(V_x\) is finite and \(S^{(r)}\subseteq S^{(r+1)}\), the sequence stabilizes after at most \(|V_x|\) iterations at
\[
S^{\star}_{x,\Delta}=\bigcup_{r=0}^{|V_x|}S^{(r)}.
\]
The set \(S^{\star}_{x,\Delta}\) is a fixed point. If \(S\) is any fixed point containing \(S_0\), induction gives \(S^{(r)}\subseteq S\) for all \(r\), hence \(S^{\star}_{x,\Delta}\subseteq S\). Thus \(S^{\star}_{x,\Delta}\) is the unique least fixed point. Define \(\mathrm{Slice}_{x}(\Delta)=S^{\star}_{x,\Delta}\). If \(C_{H,\Delta}(x)\subseteq V_x\) is the set of changed deterministic trace fields and $W_x=\sum_{z\in V_x}w_z$, set
\[
U_{H,\Delta}(x)=
\begin{cases}
W_x^{-1}\sum_{z\in C_{H,\Delta}(x)}w_z\mathbf 1\{z\notin\mathrm{Slice}_{x}(\Delta)\},&W_x>0,\\
0,&W_x=0.
\end{cases}
\]
Therefore \(U_{H,\Delta}(x)=0\) implies every changed deterministic field lies in the executed dependency closure of the audited patch support. This proves dependency confinement only; it does not identify the cause of an observed failure.

\paragraph{Proof of Theorem~\ref{thm:fresh-replay}.}
Fix a round \(r\) and condition on \(\mathcal{F}_{r-1}\). Then \(H_{r-1}\), candidate-patch set \(\mathcal{C}_r\), and all candidate definitions are fixed before replay admission. For any \(\Delta\in\mathcal{C}_r\), the replay variables \(Y_{H_{r-1},\Delta}(x_{r,i})\) are i.i.d. conditional on \(\mathcal{F}_{r-1}\), bounded in \([-C,C]\), and have conditional mean \(G_r(\Delta)\). Hence, for any \(\epsilon>0\),
\[
\Pr\!\left(G_r(\Delta)-\widehat G_r(\Delta)>\epsilon\mid\mathcal{F}_{r-1}\right)
\le \exp\!\left(-\frac{n_r\epsilon^2}{2C^2}\right).
\]
Set
\[
\epsilon_r=C\sqrt{\frac{2\log(|\mathcal{C}_r|/\delta_r^R)}{n_r}}.
\]
A union bound over candidates within round \(r\) gives
\[
\Pr\!\left(\exists\Delta\in\mathcal{C}_r:
G_r(\Delta)<\widehat G_r(\Delta)-\epsilon_r
\mid\mathcal{F}_{r-1}\right)
\le |\mathcal{C}_r|\exp\!\left(-\frac{n_r\epsilon_r^2}{2C^2}\right)
=\delta_r^R.
\]
Taking expectation over \(\mathcal{F}_{r-1}\) preserves the same unconditional bound. Since \(\sum_r\delta_r^R\le\delta_R\), a union bound over rounds shows that, with probability at least \(1-\delta_R\), every candidate in every round satisfies
\[
G_r(\Delta)\ge \widehat G_r(\Delta)-\epsilon_r.
\]
Because this event is simultaneous over the entire finite candidate class in each round, it also holds for the candidate \(\Delta_r\) selected after replay scores are observed. If CPPE accepts \(\Delta_r\), then
\[
G_r(\Delta_r)\ge \widehat G_r(\Delta_r)-\epsilon_r\ge m_r,
\]
which proves the main-text claim.

Expanding \(G_r(\Delta_r)\) gives
\[
\mathbb{E}_{x\sim\mathcal{D}}\bigl[Q(T_{H_r}^{d}(x;c_b))-Q(T_{H_{r-1}}^{d}(x;c_b))\bigr]
-\beta\mathbb{E}_{x\sim\mathcal{D}}D_{H_{r-1},\Delta_r}(x)
-\gamma\mathbb{E}_{x\sim\mathcal{D}}U_{H_{r-1},\Delta_r}(x)
\ge m_r.
\]
Rearranging yields the displayed per-round structural welfare inequality. Summing this inequality over accepted rounds gives a telescoping sum on the left:
\[
\sum_{r\in\mathcal{A}}\mathbb{E}_{x\sim\mathcal{D}}\bigl[Q(T_{H_r}^{d}(x;c_b))-Q(T_{H_{r-1}}^{d}(x;c_b))\bigr]
=\mathbb{E}_{x\sim\mathcal{D}}\bigl[Q(T_{H_K}^{d}(x;c_b))-Q(T_{H_0}^{d}(x;c_b))\bigr],
\]
where rejected rounds have \(H_r=H_{r-1}\) and contribute zero. This proves the cumulative form above. No step relates $Q$ to rendered-video utility $J$; the latter is evaluated empirically. \(\square\)

\paragraph{Proof of Proposition~\ref{prop:stochastic-gate}.}
At round $t$, let \(\widehat A_t(H)\) be the weighted empirical CPPE admission score and \(\overline A_t(H)\) its expectation. Hoeffding's inequality \citep{hoeffding1963probability,boucheron2013concentration} gives, for any fixed policy,
\[
\Pr\!\left(\left|\widehat A_t(H)-\overline A_t(H)\right|>s\right)
\le 2\exp\!\left(-\frac{2n_{V,t}s^2}{W_{\mathcal A}^2}\right).
\]
A union bound over the incumbent and $K_t$ candidates gives a simultaneous deviation bound of at most $s_t$ with probability $1-\delta_t^V$. Hence the error in a candidate-incumbent difference is at most $2s_t$, and \(\widehat{\Delta A}_t\ge m_{V,t}+2s_t\) implies \(\overline{\Delta A}_t\ge m_{V,t}\). A second union bound over rounds gives total failure probability at most $\sum_t\delta_t^V\le\delta_V$. The structural gate is checked separately. \(\square\)

\paragraph{Proof of Proposition~\ref{prop:cause-transfer}.}
Let $N_c^{pre}$ and $N_c^{post}$ be matched pre/post error masses for $c\in\bar{\mathcal C}$. Decompose their difference by errors removed and added:
\[
N_c^{pre}-N_c^{post}=r_c-a_c.
\]
Summing over all target, non-target, and \texttt{unknown} causes yields $\Delta_{\mathrm{transfer}}=\sum_c(r_c-a_c)$. Its sign is positive exactly when total removed mass exceeds total added mass. The targeted decomposition follows by separating $c^*$ from the sum. \(\square\)

\begin{lemma}[Recurrence-trigger reliability]
\label{thm:recurrence}
Theorem~\ref{thm:fresh-replay} starts after a finite candidate set has been produced; this lemma describes when local issues may enter candidate generation. Conditional on the fixed monitoring context, let \(X_1,\ldots,X_n\) be independent Bernoulli indicators of whether story-level opportunity \(l\) supports failure family \(g\). Let \(a=k/n\). Suppose non-systematic failures satisfy \(\mathbb{E}X_l\le p_0\), while a systematic policy defect satisfies \(\mathbb{E}X_l\ge p_1\), where \(p_0<a<p_1\). If CPPE triggers when \(\sum_lX_l\ge k\), then
\[
P_{\mathrm{false}}\le \exp\{-nD(a\Vert p_0)\},\qquad
P_{\mathrm{miss}}\le \exp\{-nD(a\Vert p_1)\},
\]
where \(D(a\Vert p)=a\log\frac{a}{p}+(1-a)\log\frac{1-a}{1-p}\). Consequently, if \(M\) pattern families are monitored, the family-wise false-trigger probability is at most \(M\exp\{-nD(a\Vert p_0)\}\). Choosing
\[
n\ge \max\left\{\frac{\log(M/\alpha)}{D(a\Vert p_0)},\frac{\log(1/\beta)}{D(a\Vert p_1)}\right\}
\]
controls family-wise false evolution at level \(\alpha\) and a missed systematic defect at level \(\beta\). The result makes no claim for correlated opportunities; held-out no-update calibration for such streams is an empirical procedure outside this lemma.
\end{lemma}

\paragraph{Proof of Lemma~\ref{thm:recurrence}.}
Let \(S_n=\sum_{l=1}^nX_l\). Under the non-systematic null, for any \(\lambda>0\), Markov's inequality gives
\[
\Pr(S_n\ge na)\le e^{-\lambda na}\prod_{l=1}^n\mathbb{E}e^{\lambda X_l}
\le \exp\{-n[\lambda a-\log(1-p_0+p_0e^\lambda)]\}.
\]
Optimizing over \(\lambda>0\) gives \(\lambda^\star=\log\frac{a(1-p_0)}{p_0(1-a)}\) and yields \(\Pr(S_n\ge k)\le\exp\{-nD(a\Vert p_0)\}\). A union bound over \(M\) monitored families gives the stated family-wise false-trigger bound. Under a systematic defect, apply the same Chernoff argument to the lower-tail event \(S_n<na\) with \(a<p_1\), obtaining \(\Pr(S_n<k)\le\exp\{-nD(a\Vert p_1)\}\). The sample-size condition follows by setting the family-wise false bound below \(\alpha\) and the missed-defect bound below \(\beta\). \(\square\)

\end{document}